\documentclass[acmlarge]{acmart}
\usepackage[utf8]{inputenc}
\usepackage{amsmath,amsfonts}
\usepackage{xcolor}
\usepackage{algorithm}
\usepackage{framed}
\usepackage{algpseudocode}
\algrenewcommand{\algorithmicrequire}{\textbf{Input:}}
\algrenewcommand{\algorithmicensure}{\textbf{Output:}}
\usepackage{graphicx,algorithm,algpseudocode}
\usepackage{hyperref}
\usepackage{epstopdf}
\usepackage{textcomp}
\usepackage{subcaption}
\usepackage{array}
\usepackage{footnote}
\usepackage{makecell}
\usepackage{booktabs}
\usepackage{multirow,multicol}
\usepackage{tabularx}

\makesavenoteenv{tabular}
\AtBeginDocument{%
  }

\setcopyright{acmlicensed}
\acmISBN{978-1-4503-XXXX-X/2018/06}

\begin{document}

\title[{$\mathrm{M}^3\text{PD}$ Dataset: Dual-view Photoplethysmography (PPG) Using Front-and-rear Cameras ...}]{$\mathrm{M}^3\text{PD}$ Dataset: Dual-view  Photoplethysmography (PPG) Using Front-and-rear Cameras of Smartphones in Lab and Clinical Settings}

\author{Jiankai Tang}
\email{tjk24@mails.tsinghua.edu.cn}
\authornote{Co-first author.}
\affiliation{%
  \institution{Department of Computer Science and Technology, Tsinghua University}
  \country{China}
}

\author{Tao Zhang}
\email{zt19375356@gmail.com}
\authornotemark[1]
\affiliation{%
  \institution{Department of Computer Science and Technology, Tsinghua University}
  \country{China}
}

\author{Jia Li}
\email{79333529@qq.com}
\authornotemark[1]
\affiliation{%
  \institution{Beijing Anzhen Hospital, Capital Medical University}
  \country{China}
}

\author{Yiru Zhang}
\email{yr-zhang24@mails.tsinghua.edu.cn}
\affiliation{%
  \institution{Department of Computer Science and Technology, Tsinghua University}
  \country{China}
}

\author{Mingyu Zhang}
\email{zhang-my22@mails.tsinghua.edu.cn}
\affiliation{%
  \institution{Department of Computer Science and Technology, Tsinghua University}
  \country{China}
}

\author{Kegang Wang}
\email{kegang.wang@foxmail.com}
\affiliation{%
  \institution{Department of Computer Science and Technology, Tsinghua University}
  \country{China}
}

\author{Yuming Hao}
\email{mingheguyu@163.com}
\affiliation{%
  \institution{Beijing Anzhen Hospital, Capital Medical University}
  \country{China}
}

\author{Bolin Wang}
\email{bjm79125@icloud.com}
\affiliation{%
  \institution{Beijing Anzhen Hospital, Capital Medical University}
  \country{China}
}

\author{Haiyang Li}
\email{ocean0203@163.com}
\affiliation{%
  \institution{Beijing Anzhen Hospital, Capital Medical University}
  \country{China}
}

\author{Xingyao Wang}
\email{wang_xingyao@a-star.edu.sg}
\affiliation{  
\institution{Agency for Science, Technology and Research}
  \country{Singapore}
  }
\author{Yuanchun Shi}
\email{shiyc@tsinghua.edu.cn}
\affiliation{%
  \institution{Department of Computer Science and Technology, Tsinghua University}
  \country{China}
}

\author{Yuntao Wang}
\email{yuntaowang@tsinghua.edu.cn}
\authornote{Corresponding author.}
\affiliation{%
  \institution{Department of Computer Science and Technology, Tsinghua University}
  \country{China}
}

\author{Sichong Qian}
\email{drqsc1990a@163.com}
\authornotemark[2]
\affiliation{%
  \institution{Beijing Anzhen Hospital, Capital Medical University}
  \country{China}
}

\renewcommand{\shortauthors}{Tang et al.}

\begin{abstract}
Portable physiological monitoring is essential for early detection and management of cardiovascular disease, but current methods often require specialized equipment that limits accessibility or impose impractical postures that patients cannot maintain. Video-based photoplethysmography on smartphones offers a convenient non-invasive alternative, yet it still faces reliability challenges caused by motion artifacts, lighting variations, and single-view constraints. Few studies have demonstrated that this technology can be reliably applied to physiological monitoring of cardiovascular patients, and no widely used open datasets exist for researchers to examine its cross-device accuracy. To address these limitations, we introduce the $\mathrm{M}^3\text{PD}$ dataset—the first publicly available dual-view mobile photoplethysmography dataset—comprising synchronized facial and fingertip videos captured simultaneously via front and rear smartphone cameras from 60 participants (including 47 cardiovascular patients). Building on this dual-view setting, we further propose the $\mathrm{F}^3\text{Mamba}$, which fuses the facial and fingertip views through Mamba-based temporal modeling. The model reduces heart-rate error by 21.9--30.2\% over existing single-view baselines while showing enhanced robustness across challenging real-world scenarios. Data and code are released at \url{https://github.com/Health-HCI-Group/F3Mamba/tree/main}.
\end{abstract}



\maketitle

\section{Introduction}
Portable health monitoring is important for everyday well-being, with heart rate (HR) serving as a key physiological indicator for cardiovascular health assessment~\cite{ma2025noncontact,bian2024ubihr}. Traditional HR measurement methods rely on specialized medical devices such as electrocardiography (ECG) or photoplethysmography (PPG) sensors that are often inconvenient to carry and impractical for continuous monitoring~\cite{tang2023mmpd}. This creates a critical gap in surveillance particularly for patients with arrhythmias, hypertension, and coronary artery disease, where cardiovascular events can occur suddenly and unpredictably~\cite{liu2022vidaf}. By the time patients reach medical facilities, transient cardiac abnormalities may have already normalized, making timely diagnosis and appropriate treatment decisions more challenging~\cite{chong2013arrhythmia}.

Recent studies in ubiquitous and mobile computing have shown that smartphones themselves can be turned into physiological sensors: inertial-sensor–based methods can reconstruct ECG-like signals or estimate HR and heart rate variability (HRV) from accelerometers and gyroscopes~\cite{wang2023ecg,Mohamed2017heartsense}, and acoustic sensing can even pick up heartbeat-induced chest motion using commodity smart speakers~\cite{zhang2020speaker}. These efforts clearly validate the idea of phone-centric health monitoring. However, to reach their reported accuracy, most of these systems still assume static or semi-static postures, stable phone–body placement, and low ambient noise. Once the phone is truly handheld, the user is elderly, or the scene contains natural head/hand movements, the inertial or acoustic channels are easily flooded by motion artifacts. This suggests that we need a more stable sensing modality that works \emph{with} natural smartphone use rather than against it.

\begin{figure*}[t] 
  \centering
  \includegraphics[width=\textwidth]{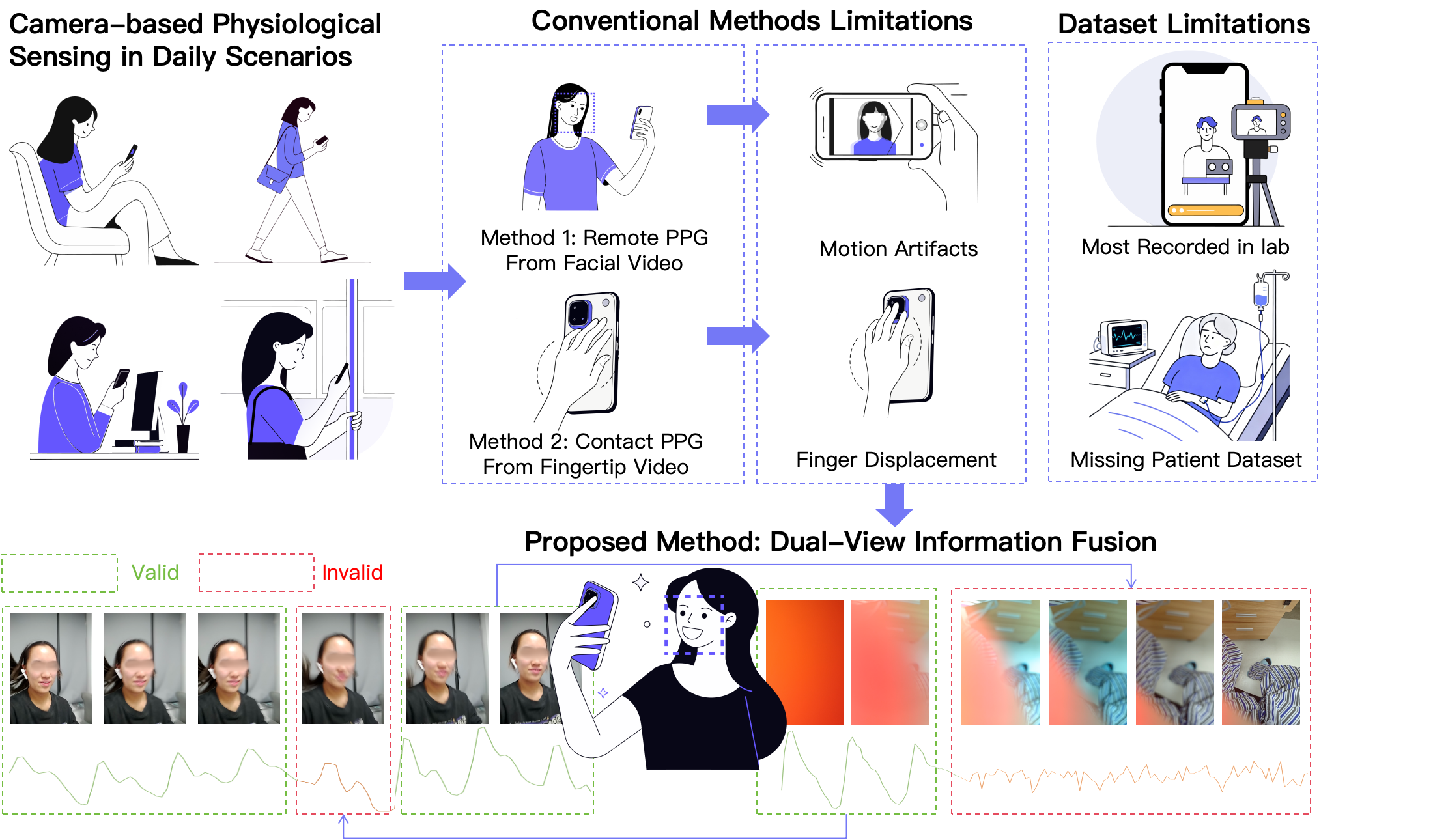}
  \caption{\textbf{Fusion of video-based physiological sensing.} Video-based physiological sensing faces challenges from motion artifacts, lighting variations, and position instability. Traditional approaches rely on single views (facial or fingertip), limiting robustness. Our dual-position fusion method integrates signals from both front camera (facial) and rear camera (fingertip) videos. The $\mathrm{F}^3\text{Mamba}$ framework leverages this dual-view approach to enhance algorithm robustness and accuracy in heart rate estimation across real-world scenarios.}
  \label{fig:fusion} 
\end{figure*}

Video-based physiological sensing has emerged as a promising solution to address these challenges, enabling non-invasive extraction of vital signs from facial~\cite{tang2024camera} or fingertip regions~\cite{hoffman2022smartphone} without requiring specialized equipment. With recent advancements in smartphone camera technology and computational capabilities, an increasing number of studies have explored mobile phone-based remote physiological sensing applications that can be integrated into everyday life~\cite{wang2024plug,liu2023rppg}. Unlike traditional fixed-camera approaches used in clinical settings~\cite{liu2022vidaf}, smartphone cameras offer superior portability and accessibility~\cite{liu2022MobilePhys}, making physiological monitoring feasible across diverse environments and populations.

However, extracting physiological signals from videos using smartphone cameras still faces reliability challenges in healthcare settings. Smartphones are usually operated in a handheld manner, which introduces motion artifacts and jitter that can compromise physiological signal quality~\cite{liu2024spiking}. These challenges are even more pronounced for elderly users and cardiovascular disease (CVD) patients, who may find it difficult to maintain a stable posture or fixed device position during measurement.

Existing methods therefore often resort to stationary or constrained setups—such as tripod-mounted smartphones for facial monitoring~\cite{shoushan2021noncontact,viplhr} or requiring the hand to be placed on a stable surface~\cite{aziz2021automated,bian2024ubihr}—but such assumptions do not reflect real-world mobile usage. Variations in camera-to-subject distance, viewing angle, and ambient illumination further complicate video-based physiological signal extraction~\cite{wang2024plug}. These limitations are particularly problematic in clinical or pre-clinical screening scenarios, where measurement accuracy directly affects diagnostic value. Recent cross-dataset evaluations~\cite{liu2023rppg} on the MMPD smartphone video dataset~\cite{tang2023mmpd} have shown that even advanced algorithms can yield heart-rate estimation errors exceeding 10 beats per minute (BPM), which falls short of the precision required for many cardiovascular assessments.

A key reason for this performance gap is that current smartphone-based physiological sensing pipelines typically rely on a single view (for example, facial~\cite{shoushan2021noncontact} or fingertip~\cite{aziz2021automated}) and thus fail to exploit the complementary information available from multiple sensing sites. This is a critical limitation for CVD patients, whose peripheral perfusion and physiological waveforms may fluctuate across time and body locations, making one view (e.g., face) unreliable while another view (e.g., fingertip) still contains usable pulsatile components. Yet, to date, few studies have systematically explored the potential of simultaneous front–rear smartphone camera recording to improve robustness. Early work such as MobilePhys~\cite{liu2022MobilePhys} showed that rear-camera signals can enhance performance but required subject-specific retraining, which restricts deployment. The most relevant study used two USB cameras on only 10 subjects~\cite{liu2024summit}, suggesting that multi-position video input can improve heart-rate estimation, but it relied on external cameras rather than on-board dual cameras, limiting its practicality for everyday mobile health monitoring.

To address these limitations in smartphone-based physiological monitoring—especially the lack of dual-view mobile video data for CVD patients—we introduce the \textbf{M}ulti-view \textbf{M}ulti-scenario \textbf{M}obile \textbf{P}hysiology \textbf{D}ataset (\textbf{$\mathrm{M}^3 \text{PD}$}). As illustrated in Figure~\ref{fig:fusion}, our dual-view fusion approach integrates signals from both front camera (facial) and rear camera (fingertip) videos to enhance robustness against motion artifacts, lighting variations, and position instability. The $\mathrm{M}^3 \text{PD}$ dataset is, to the best of our knowledge, the first publicly available smartphone dual-view physiological sensing dataset that explicitly targets handheld and clinically relevant scenarios. It contains synchronized facial and fingertip videos recorded simultaneously by front and rear smartphone cameras in both lab and clinical settings, together with clinical-grade physiological measurements including PPG waveforms, blood oxygen (SpO2), and blood pressure (BP). The dataset comprises recordings from 60 subjects, among which 47 are CVD patients, enabling validation in both laboratory (n=13) and clinical (n=47) environments. Building on this dual-view setting, we further propose the \textbf{F}acial-\textbf{F}ingertip \textbf{F}usion \textbf{Mamba} (\textbf{$\mathrm{F}^3\text{Mamba}$}) framework to integrate complementary physiological information from dual-view streams. $\mathrm{F}^3\text{Mamba}$ dynamically updates state representations across views and performs temporal fusion through a Fusion Mamba (F-Mamba) architecture, yielding more reliable heart-rate estimates even when one view is corrupted by motion, low perfusion, or lighting artifacts.

\begin{table*}[htp]
  \caption{\textbf{Datasets Comparison.} Details of wide-use video physiological sensing datasets.}
  \centering
  \begin{tabular}{cccccc}
  \toprule[1.5pt]
  Dataset & Scenarios & Subjects & Camera & Position & Vitals   \\
  \hline
  \hline
  PURE~\cite{stricker2014non} & Lab & 10 & eco274CVGE & Face & PPG/SpO$_2$  \\
  UBFC-rPPG~\cite{ubfcrppg} & Lab& 42 & Logitech C920 & Face & PPG   \\
  Oximetry~\cite{hoffman2022smartphone} & Lab & 6 & Google Nexus 6P & Finger & SpO$_2$ \\
  MMPD~\cite{tang2023mmpd} & Lab & 33 & Galaxy S22 Ultra & Face & PPG   \\
  RLAP~\cite{wang2024camera} & Lab & 58 & Logitech C930c & Face & PPG \\
  SUMS~\cite{liu2024summit} & Lab & 10 & Logitech C922 & Face+Finger & PPG/SpO$_2$/RR  \\
  LADH~\cite{ma2025noncontact} & Lab & 21 & Logitech C922 & Face(RGB+IR)  & PPG/SpO$_2$/RR   \\
  \hline
  \multirow{2}{*}{$\mathrm{M}^3 \text{PD}${(\textbf{Ours})}} & Lab & 13& OPPO A52 & Face+Finger & PPG/SpO$_2$/RR/BP  \\
  & Clinic & 47& XiaoMi 14 & Face+Finger & PPG/SpO$_2$/RR/BP\\
  \bottomrule[1.5pt]
  \end{tabular}%
  \label{tab: dataset}
  \end{table*}

The main contributions of this paper are:

\begin{itemize}
  \item We present $\mathrm{M}^3 \text{PD}$, the first dual-view smartphone dataset that records front-camera (face) and rear-camera (fingertip) videos from \emph{CVD patients} (n=47) and healthy subjects (n=13). This resource addresses realistic handheld challenges faced in point-of-care cardiovascular monitoring, including motion artifacts and unstable handling, particularly among elderly CVD patients.
  \item We develop the $\mathrm{F}^3 \text{Mamba}$ framework, which explicitly models facial and fingertip videos as two complementary views and fuses them through view-specific Temporal Difference Mamba (TD-Mamba) blocks and a cross-view F-Mamba module. This design enables dynamic state propagation across views, so that the system can rely on the more reliable stream when one view is degraded (e.g., face under motion, fingertip under low perfusion).
  \item On $\mathrm{M}^3 \text{PD}$, our fusion strategy reduces heart-rate estimation error by \textbf{21.9--30.2\%} compared with state-of-the-art single-view baselines, and the gains hold on both controlled lab and cadiogy clinic, demonstrating that dual-view fusion is not only algorithmically beneficial but also \emph{clinically relevant} for telemedicine applications.
\end{itemize}

\section{Related Works}
\subsection{Video Physiology Dataset}

The development of multi-view datasets has been essential for advancing remote photoplethysmography (rPPG) in patient care applications. These datasets typically include synchronized recordings of facial videos alongside cardiovascular measurements such as PPG waveforms and heart rate. The PURE dataset \cite{stricker2014non} represents one of the earliest contributions, featuring facial videos captured under laboratory conditions with corresponding PPG and SpO$_2$ measurements from pulse oximeters. The UBFC-rPPG dataset \cite{ubfcrppg} expanded this approach with a larger participant pool using USB webcams. The Oximetry dataset~\cite{hoffman2022smartphone} shifted focus to fingertip videos from rear smartphone cameras to predict SpO$_2$ levels during controlled oxygen desaturation protocols. The MMPD dataset \cite{tang2023mmpd} addressed variety by including multiple skin tones, lighting conditions, and movement patterns using fixed-position smartphones. The RLAP dataset \cite{wang2024camera} further improved data quality with standardized recording protocols across various scenarios.

Recent datasets have widened the scope of physiological monitoring beyond basic heart rate detection, reflecting the growing potential of non-contact sensing for thorough cardiovascular assessment. The SUMS dataset \cite{liu2024summit} introduced dual-view collection (face and fingertip) specifically designed for monitoring hypoxic conditions in highland, with oxygen saturation levels as low as 90\%—medically relevant for patients with respiratory disorders. The LADH dataset \cite{ma2025noncontact} advanced monitoring capabilities by incorporating infrared facial recordings that maintain accuracy despite face coverings, enabling continuous physiological tracking over extended periods. These developments represent a natural progression toward integrated monitoring systems capable of assessing multiple cardiovascular parameters simultaneously, supporting more complete patient evaluation in both clinical and home settings.

However, existing datasets fail to address a key challenge in applying rPPG technology to everyday medical monitoring—the natural movement artifacts introduced during handheld smartphone use. Most current datasets rely on stationary or tripod-mounted cameras in controlled environments, creating a notable gap between laboratory performance and real-world clinical utility. While some datasets like VIPL~\cite{viplhr} and MMPD~\cite{tang2023mmpd} have incorporated limited handheld scenarios, they mainly feature brief, stable recordings that do not reflect typical patient usage patterns. This limitation is especially important for cardiovascular monitoring in elderly patients and those with limited dexterity, who often struggle to maintain stable device positioning during measurement. 

To address this critical gap in clinical usefulness, we developed the \textbf{M}ulti-view \textbf{M}ulti-scenario \textbf{M}obile \textbf{P}hysiology \textbf{D}ataset ($\mathrm{M}^3 \text{PD}$), which captures both facial and fingertip videos using handheld smartphones in both laboratory and clinic environments. By intentionally including the natural movement patterns observed in everyday clinical practice, this dataset enables the development of more robust algorithms that can maintain accurate cardiovascular measurements despite the variable recording conditions encountered in real-world patient monitoring.

\subsection{rPPG Algorithms}

rPPG algorithms have evolved significantly, leveraging advancements in computer vision and signal processing to extract physiological signals from video data. These algorithms can be broadly categorized into two main approaches: traditional unsupervised methods and supervised deep-learning methods.

Traditional rPPG methods primarily rely on signal processing and color space analysis to extract physiological signals from facial videos. Verkruysse et al. \cite{verkruysse2008remote} first demonstrated that ambient light and the green channel of RGB video can be used for remote plethysmographic imaging, laying the foundation for subsequent research. Poh et al. \cite{poh2010advancements} introduced Independent Component Analysis (ICA) to separate the pulse signal from noise in webcam videos, improving robustness to environmental variations. De Haan et al. \cite{de2013robust} proposed the CHROM method, which leverages chrominance-based signal processing to enhance pulse rate estimation accuracy under varying lighting conditions. To address motion artifacts, Wang et al. \cite{wang2016algorithmic} introduced the Plane-Orthogonal-to-Skin (POS) algorithm, which formulates rPPG extraction as a projection problem in color space, significantly enhancing signal quality. Álvarez et al. \cite{
} hysroposed Face2PPG, an unsupervised pipeline for extracting blood volume pulse signals from facial videos, further advancing the field toward practical, real-world applications.

However, these traditional methods often rely on restrictive assumptions about stable illumination and minimal motion, limiting their effectiveness in unconstrained real-world environments. To address these limitations, recent research has shifted toward deep learning approaches for enhanced robustness and generalization in variable conditions.

Early deep learning contributions include DeepPhys \cite{chen2018deepphys} and PhysNet \cite{yu2019remote}, which pioneered end-to-end neural networks for video-based vital sign measurement with improved spatio-temporal feature learning. Subsequently, Transformer architectures were adapted to rPPG research, leveraging their ability to model long-range dependencies in temporal data. PhysFormer \cite{yu2022physformer} and RhythmFormer \cite{zou2025rhythmformer} demonstrated significant accuracy improvements through effective capture of complex temporal and rhythmic patterns.

More recent advances include PhysMamba \cite{luo2024physmamba}, which combines TD-Mamba blocks with a dual-stream SlowFast architecture to enhance local dynamics while maintaining long-range context for robust heart rate estimation. Similarly, MaKAN-Mixer \cite{zhang2025makan} integrates Eulerian Video Magnification with Temporal Shift Module Amplification to enhance subtle physiological signals, while employing a Mamba-KAN Fusion Module for efficient temporal modeling and channel mixing.

Despite these advances, most existing rPPG algorithms are still designed and evaluated in a \emph{single-view} setting: they operate either on facial videos (remote, non-contact) or on fingertip videos (contact, rear camera + flash), and their feature modeling is mainly based on (i) color-space redundancy and (ii) periodic temporal patterns. Because these principles hold for both facial and fingertip recordings, we include representative unsupervised and deep-learning baselines in our evaluation on $\mathrm{M}^3\text{PD}$. Yet, to the best of our knowledge, no prior work explicitly exploits \emph{both} synchronized views from the same smartphone to perform complementary physiological estimation, especially for low-perfusion or arrhythmic cardiovascular patients. To fill this gap, we propose $\mathrm{F}^3\text{Mamba}$, which performs cross-view fusion over temporally aligned facial and fingertip streams, and we benchmark it against these widely used single-view baselines on our dual-view dataset to quantify the benefit of multi-view modeling.

\begin{figure*}[t] 
  \centering
  \includegraphics[width=\textwidth]{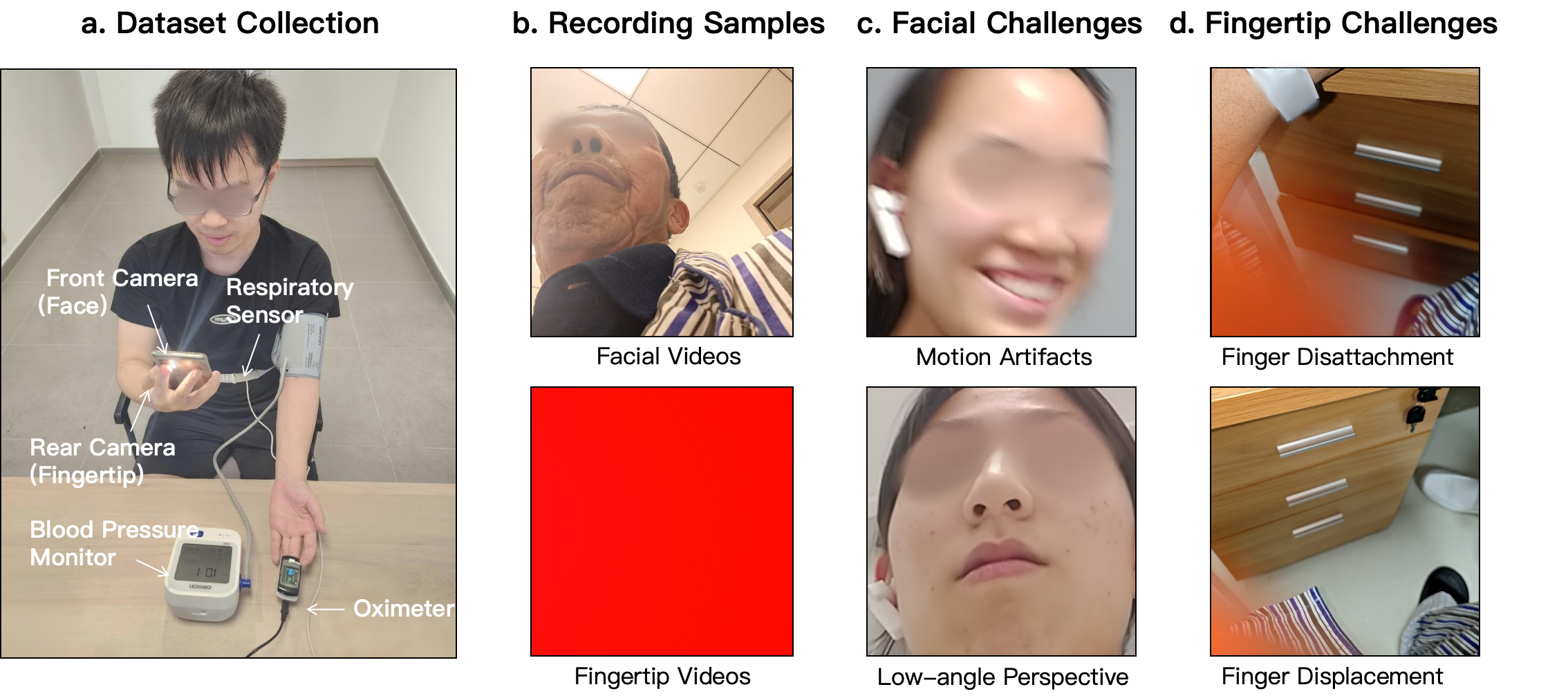}
  \caption{\textbf{Data collection setup and real-world challenges in dual-view mobile rPPG.} (a) Synchronized data acquisition system capturing facial and fingertip videos simultaneously via front and rear smartphone cameras, with concurrent physiological measurements including respiratory sensor, blood pressure monitor, and pulse oximeter. (b) Representative recording samples showing facial videos from elderly cardiovascular patients and fingertip videos with characteristic red appearance from rear camera flash. (c) Facial video challenges during handheld recording: motion artifacts from natural head movements and low-angle perspective distortions common in patient self-monitoring. (d) Fingertip video challenges: finger disattachment from camera surface and lateral finger displacement, particularly prevalent among elderly users with limited dexterity.}
  \label{fig:collection} 
\end{figure*}

\subsection{Mamba Fusion}
Recent advances in multimodal fusion have driven significant progress across diverse domains including medical imaging\cite{li2025bsafusion}, autonomous driving \cite{peng2025moral}, remote sensing\cite{peng2023u2net}, and human-computer interaction\cite{fu2024cross}. Traditional approaches to fusion include early/late integration strategies \cite{tsai2019multimodal,simonyan2014two}, hybrid feature combination\cite{choi2024fusion}, and attention-based cross-modal interaction\cite{yin2016abcnn,liang2023cross}. While these methods can combine information from different modalities, they often struggle with computational efficiency and comprehensive fusion when processing high-dimensional data.

The emergence of Mamba, a state-space model (SSM) based architecture, offers a promising solution by maintaining linear time complexity while achieving better scalability than Transformers. Vision Mamba \cite{zhu2024vision} first applied SSMs to visual tasks, inspiring subsequent multimodal fusion frameworks. Several recent studies have tailored Mamba for multimodal applications with notable success. Xie et al. \cite{xie2025fusionmamba} developed a cross-modal fusion Mamba specifically designed for detailed interaction between modalities. Dong et al. \cite{dong2025hybridfusion} introduced HFMamba, a lightweight network that uses dual Mamba branches to extract and hierarchically fuse complementary features from different perspectives.

In the remote sensing domain, researchers have adapted Mamba for specialized fusion needs. Peng et al. \cite{peng2024fusionmamba} created a dual-input Mamba block that dynamically combines spatial and spectral features through an interactive SSM update mechanism. Similarly, Cao et al. \cite{caom3amba} designed a cross-attention module (Cross-SS2D) that efficiently exchanges information between multimodal data by using complementary inputs to refine SSM parameters.


While these approaches show Mamba's effectiveness in fusing homogeneous views (such as spatial and spectral features from the same region), they typically rely on strong inherent correlations between the data sources. Our task presents a different challenge: we must integrate physiological signals from two distinct locations (face and fingertip) that lack direct spatial correspondence. This separation introduces unique difficulties in developing efficient state-space model interactions capable of accurately estimating heart rate from these complementary yet weakly aligned views. Unlike previous approaches designed for closely related inputs, our method must bridge the physiological gap between these different vascular regions.

\section{Dataset}

In this section, we introduce the $\mathrm{M}^3 \text{PD}$ dataset, which is the first publicly available dual-view physiological sensing dataset captured using handheld smartphones. We start by describing the data collection system in \autoref{sec:system}, followed by details of the lab dataset in \autoref{sec:lab_dataset} and the Clinic dataset in \autoref{sec:clinic_dataset}. We summarize the dataset characteristics in \autoref{tab: dataset}. 


\subsection{Collection System}
\label{sec:system}
This section describes the synchronized multi-modal data acquisition system used to collect the $\mathrm{M}^3 \text{PD}$ dataset, including hardware components, software applications, and data synchronization methods.

\subsubsection{Hardware}
\begin{figure}[H]
  \centering
  \includegraphics[width=0.6\linewidth]{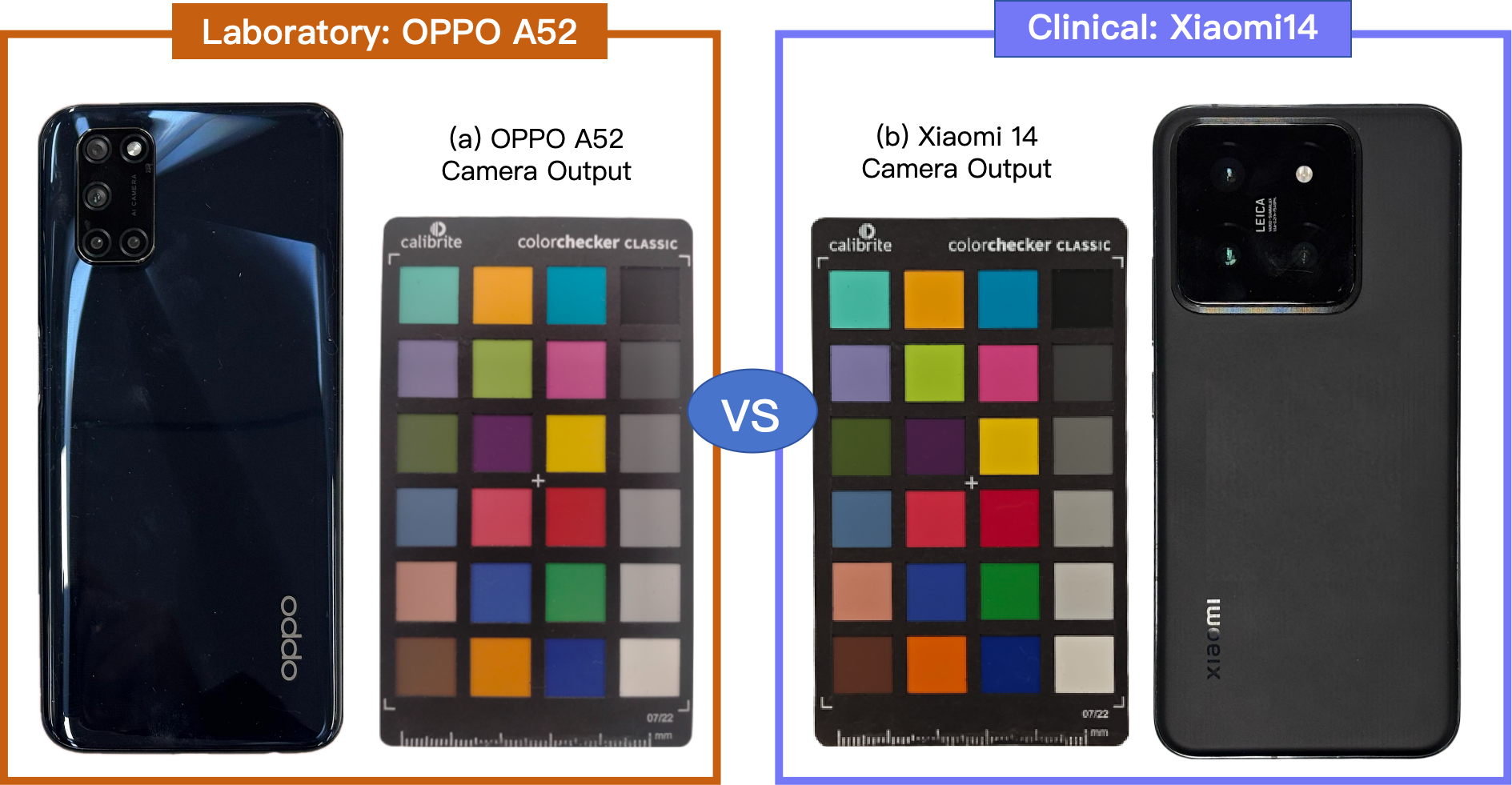}
  \caption{\textbf{Camera color reproduction variability across devices and environments.} Comparison of ColorChecker Classic captured by (a) Xiaomi 14 in clinical settings and (b) OPPO A52 in laboratory settings. The distinct color reproductions reflect differences in camera sensors and image signal processors (ISPs) between devices, demonstrating the hardware variability that algorithms must handle for robust cross-device generalization in real-world mobile health monitoring applications.}
  \label{fig:color_variability}
\end{figure}

The hardware setup comprises a central Windows computer, an Android smartphone, and several medical-grade sensors. Two different smartphone models were used to capture data across the two study environments: an OPPO A52 in the lab and a Xiaomi 14 in the clinic. These devices feature distinct camera sensors and image signal processors (ISPs), resulting in inherent differences in color reproduction, noise characteristics, and video processing pipelines. To characterize these variations, we recorded a standard color chart with both phones, confirming that the captured data reflects the diversity of consumer devices, as illustrated in \autoref{fig:color_variability}. This hardware variability is crucial for developing and validating rPPG algorithms that can generalize across different devices in real-world settings.

For ground-truth physiological measurements, we used a CMS50E pulse oximeter to record PPG waveforms at 20 Hz and SpO$_2$ at 1 Hz, an HKH11C respiratory belt for breathing waveforms at 50 Hz, and an OMRON U726J automated cuff for blood pressure readings. Ground-truth devices are demonstrated in \autoref{fig:collection}(a).

\subsubsection{Software}
\begin{figure*}[!ht]
  \centering
  \includegraphics[width=0.9\textwidth]{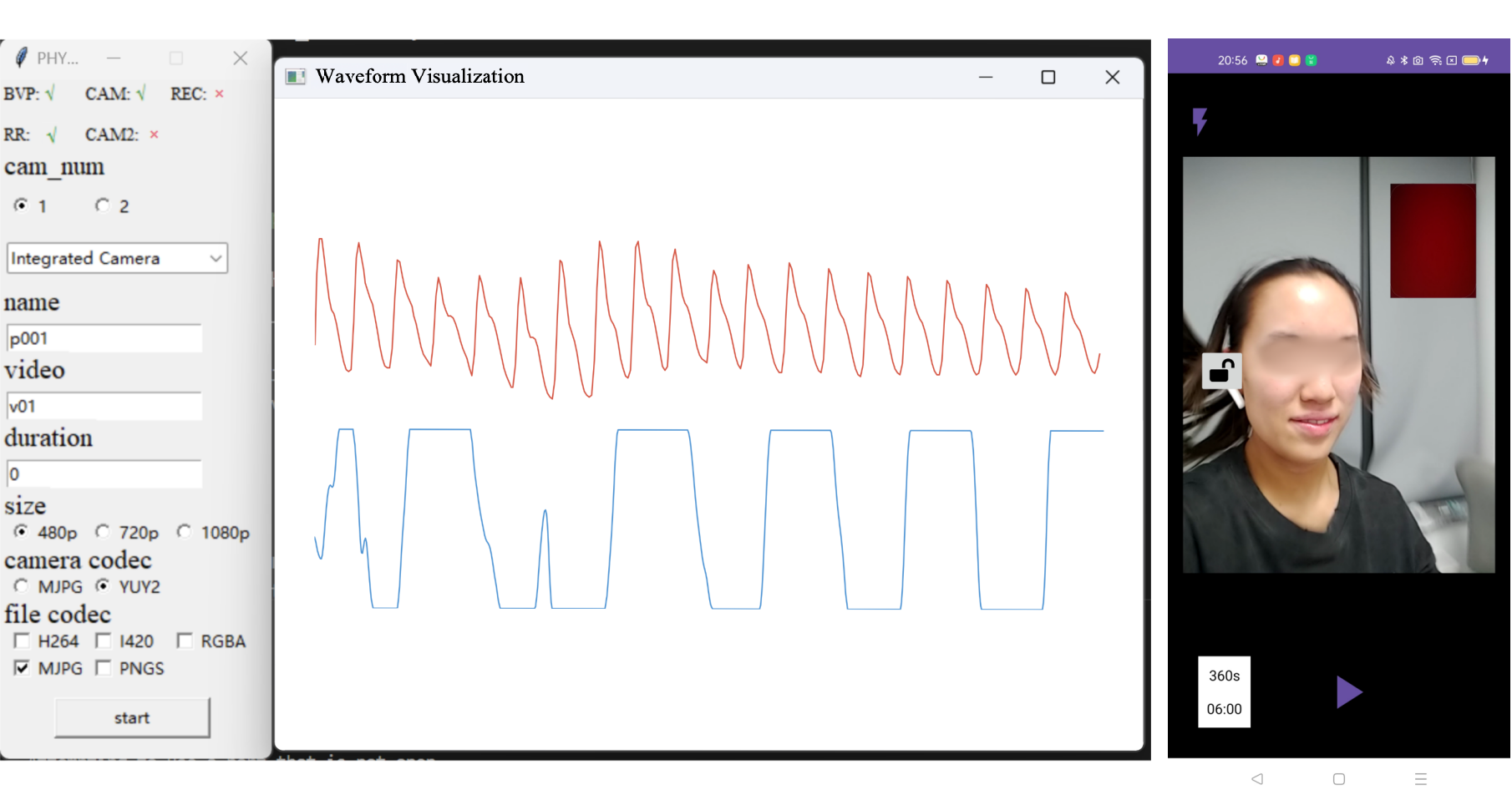}
  \caption{\textbf{Synchronized multi-modal data acquisition system.} The system interface displays real-time physiological waveforms including blood volume pulse (BVP, top) and respiratory rate (RR, bottom) signals synchronized with simultaneous dual-view smartphone recording. The right panel shows the mobile application interface capturing both facial (front camera) and fingertip (rear camera) videos with real-time preview and recording controls.}
  \Description{Screenshot showing a synchronized data acquisition system with three main components: left panel displaying real-time BVP and respiratory waveforms, center control panel with recording settings, and right panel showing the mobile app interface with dual camera views for facial and fingertip video capture.}
  \label{fig:data_system}
\end{figure*}

As illustrated in \autoref{fig:data_system}, our system includes two custom software applications: a data acquisition platform on the Windows computer and a video recording application on the Android smartphone. 

Inspired by PhysRecorder~\cite{wang2024camera}, the Windows platform provides a centralized interface for initiating and monitoring data streams from the connected medical sensors. The Android application simultaneously records video from the front (face) and rear (fingertip) cameras at a resolution of 1280x720 and a frame rate of 30 fps, embedding precise timestamps for each frame.

\subsubsection{Data Synchronization}
Ensuring precise temporal alignment between video streams and physiological signals is critical for rPPG research. Our system achieves this through a multi-level synchronization strategy. The smartphone and the Windows computer are synchronized to the same Network Time Protocol (NTP) server, establishing a common time reference. The medical sensors are connected to the computer via serial ports, which allows the data acquisition software to record incoming physiological data with millisecond-precision timestamps relative to the system clock. During data processing, all data streams—facial video, fingertip video, PPG, and respiration—are aligned using their respective timestamps, guaranteeing accurate correspondence between video frames and ground-truth physiological events.

\subsection{Controlled Lab Dataset}
\label{sec:lab_dataset}

\subsubsection{Participants}
We recruited 13 healthy adults (6 male, 7 female; age 18--30 years, mean $21.38 \pm 3.78$) to participate in the laboratory study. All participants provided written informed consent before the experiment. The study protocol was reviewed and approved by the local institutional review board (IRB) of the authors' affiliation. This subset is intended to provide a clean, well-controlled source domain that can be contrasted with the more challenging patient recordings in \autoref{sec:clinic_dataset}.

\subsubsection{Data Collection Procedure}
Data were recorded using an OPPO~A52 smartphone configured to capture \emph{simultaneous} dual-view videos: the front camera recorded the participant’s face, while the rear camera (with LED flash) recorded a fingertip video under contact illumination. The phone was time-synchronized with the Windows-based acquisition computer described in \autoref{sec:system}, which received physiological signals from a CMS50E pulse oximeter (PPG waveform at 20~Hz, SpO$_2$ and HR at 1~Hz), an HKH11C respiratory belt (50~Hz), and an OMRON U726J automated cuff (pre/post blood-pressure readings). All videos were stored at 1280$\times$720 resolution and 30~fps, with per-frame timestamps embedded to support alignment with the physiological streams.

The laboratory protocol was designed to emulate typical mobile cardiovascular self-monitoring behaviors and consisted of five phases (see \autoref{fig:experiment_protocol}):  

\begin{figure}[H]
  \centering
  \includegraphics[width=\linewidth]{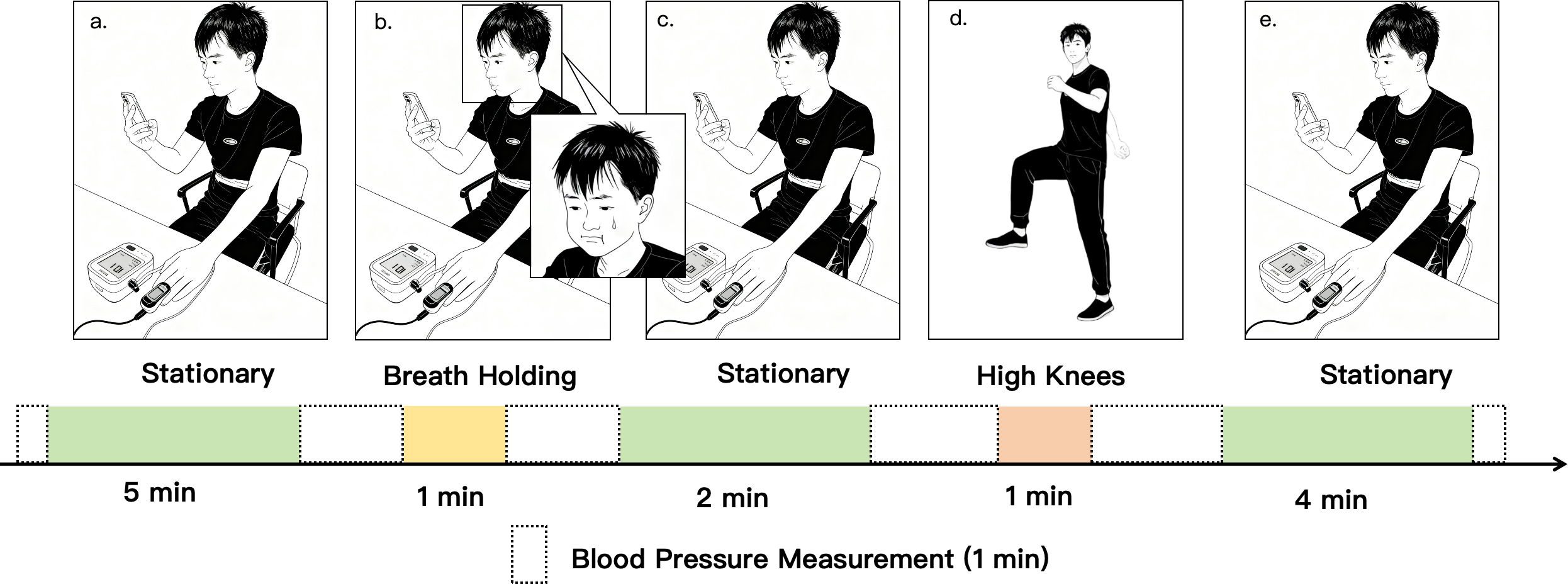}
  \caption{\textbf{Experimental protocol for data collection.} The protocol consists of five phases designed to simulate real-world cardiovascular monitoring scenarios: baseline resting state (5 min), breath-holding for autonomic response testing (1 min), recovery period (2 min), high leg lifts for exertional heart rate changes (1 min), and final recovery phase (4 min). Blood pressure measurements were taken during the breath-holding phase to capture comprehensive cardiovascular parameters.}
  \label{fig:experiment_protocol}
\end{figure}

(i) 5~min seated rest (baseline),  
(ii) 1~min breath-holding to elicit an autonomic response,  
(iii) 2~min seated recovery,  
(iv) 1~min high leg lifts to induce exertional heart-rate changes, and  
(v) 4~min final seated rest for post-exertion monitoring.  
Each session lasted about 15~min in total, yielding roughly 13~min of effective dual-view recording per participant. During the high leg-lift phase, strong body motion caused noticeable corruption in the contact oximeter reference; therefore, this phase is \emph{not} used in our quantitative benchmarks, but the corresponding facial and fingertip videos are kept in the released dataset to support research on motion-robust rPPG, view completion, and quality assessment.

\subsubsection{Dataset Characteristics and Challenges}
Although collected in a laboratory, the recordings still exhibit common real-world artifacts (see \autoref{fig:collection}):  
(1) video jitter introduced by handheld or slightly moving smartphones;  
(2) variations in facial pose, distance, and partial face visibility when participants adjust their sitting posture; and  
(3) fingertip displacement or partial detachment from the rear camera surface, especially during transitions between phases.  
These effects reduce the effective pulsatile component in both facial and fingertip videos and make the lab subset more representative of actual mobile health usage than fully constrained datasets in \autoref{tab: dataset}. By releasing both the clean resting segments and the more challenging motion/transitional segments, $\mathrm{M}^3\text{PD}$ allows researchers to evaluate best-case rPPG accuracy, to test robustness to short motion bursts, and—most importantly for our work—to study whether dual-view fusion can compensate for temporary signal degradation in either view.

\subsection{Clinic Dataset}
\label{sec:clinic_dataset}

\subsubsection{Participants}
We collected the clinical subset from 47 outpatients with documented cardiovascular conditions (30 male, 17 female; age 24--78 years, mean $60.3 \pm 10.6$). The cohort covered common diagnoses such as coronary artery disease, chronic heart failure, and atrial fibrillation. These conditions are characterized by unstable hemodynamics, rhythm irregularity, or reduced peripheral perfusion, all of which are known to make camera-based rPPG less reliable. All participants signed informed consent, and the study was approved by the institutional review board of the collaborating clinical site.

\subsubsection{Dataset Collection Procedure}
Recordings were conducted in a seated position in a real clinical environment. Each participant was asked to hold a Xiaomi~14 smartphone in front of them and look toward the screen while the device \emph{simultaneously} captured (i) a facial video from the front camera and (ii) a fingertip video from the rear camera with flash. Importantly, we did \emph{not} constrain how participants gripped the phone or how stably they maintained the device and head pose; small hand tremors, low-angle views, and micro-adjustments of the phone were allowed because they frequently occur in outpatient self-check scenarios. Each recording lasted about 30~s, which fits into the clinical workflow and is sufficient for heart-rate estimation from both views. The smartphone was time-aligned with the Windows-based acquisition system described in \autoref{sec:system}, so that dual-view videos and physiological references (CMS50E PPG/HR/SpO$_2$, and spot BP when available) share a common timestamp. All videos were stored at 1280$\times$720 resolution and 30~fps, identical to the lab subset to enable joint training and cross-subset evaluation.

\subsubsection{Dataset Characteristics and Challenges}
As shown in \autoref{fig:collection}, the clinical recordings expose two characteristic sources of difficulty:

(1) \textbf{Physiological variability.} Many of the enrolled patients presented arrhythmias (e.g., AF), lower pulsatile amplitude, or disease-related changes in peripheral circulation. In patients with cardiovascular disease, hemodynamic alterations and impaired peripheral perfusion often lead to weak or irregular pulsations at distal sites (e.g., fingertip and face). Because PPG relies on detecting small blood-volume changes in the microvascular bed, such low-perfusion or low-pulsatility conditions reduce the signal-to-noise ratio and make beat detection less reliable, especially for camera-based rPPG. This mechanism has been reported to cause pulse underestimation and increased error in the presence of arrhythmias or peripheral vascular dysfunction, which is exactly the population we capture here. As a result, even when the medical reference reports a stable or elevated heart rate, the facial or fingertip optical signals in our dataset may show intermittently missing pulses.

(2) \textbf{Handling variability.} Because participants were not forced to fix their posture or grip, natural hand tremors, brief fingertip detachment from the rear camera, and changes in facial angle are commonly observed. These factors introduce frame-to-frame motion and view changes that are rarely seen in controlled datasets.

Unlike the lab subset, \emph{all} clinical recordings, including those with these real-world artifacts, are included in our benchmark experiments. This design choice is intentional: mobile health systems for cardiovascular patients must operate under exactly these conditions, so we preserve them to let researchers evaluate robustness, view-level failure handling, and cross-view fusion on a realistic patient population.

\begin{table}[t]
    \centering
    \caption{Sample counts for Lab and Clinic subsets.}
    \label{tab:data_stats}
    \begin{tabular}{lrrrrrrr}
        \toprule
      Scenario  & Facial Frames & Fingertip Frames & BVP & HR& RESP & SpO2 & BP \\
        \midrule
        Lab    & 366{,}501 & 366{,}718 & 376{,}942 & 14{,}369 & 959{,}573 & 14{,}369 & 52 \\
        Clinic & 49{,}242  & 49{,}014  & 28{,}585  & 1{,}412  & --        & 1{,}412  & 47 \\
        \bottomrule
    \end{tabular}
\end{table}

\subsection{Dataset Structure}
\begin{figure*}[htbp] 
  \centering
  \includegraphics[width=\textwidth]{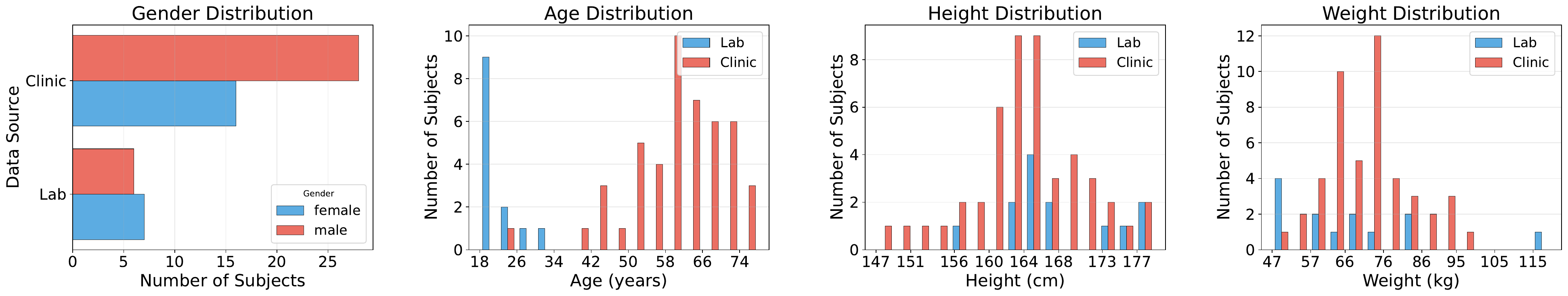}
  \caption{\textbf{Demographic distribution of participants in the Lab and Clinic datasets.} From left to right: (a) Gender distribution, (b) Age distribution, (c) Height distribution, and (d) Weight distribution. The Clinic dataset includes a higher proportion of older participants, reflecting the target population for cardiovascular monitoring.}
  \label{fig:subject_distribution} 
\end{figure*}

The $\mathrm{M}^3 \text{PD}$ dataset is organized into two main subsets: the Lab dataset and the Clinic dataset. Each subset contains synchronized facial and fingertip videos along with corresponding physiological measurements. The data sample points statistics are shown in ~\autoref{tab:data_stats}. Demographic distribution of the dataset is illustrated in ~\autoref{fig:subject_distribution}. The dataset structure is as follows:

\begin{small}
\begin{framed}
\begin{verbatim}
Dataset/
|-- 001/                         # subject ID
|   |-- dual_camera_session_20250115_103012/
|   |   |-- front_camera_20250115_103012.mp4     # facial view (front camera)
|   |   |-- back_camera_20250115_103012.mp4      # fingertip view (rear camera)
|   |   |-- front_camera_meta_20250115_103012.txt
|   |   `-- back_camera_meta_20250115_103012.txt
|   `-- v01/              # synchronized physiological labels
|       |-- BVP.csv
|       |-- HR.csv
|       |-- RR.csv
|       |-- SpO2.csv
|       `-- frames_timestamp.csv  # mapping video frame -> signal time
|-- 002/
|   `-- ...
|-- ...
\end{verbatim}
\end{framed}
\end{small}

\subsection{Data Release and Access Policy}
The $\mathrm{M}^3\text{PD}$ dataset contains two types of sensitive information: (i) full facial videos that can reveal the participant’s identity, and (ii) clinical physiological measurements from cardiovascular patients. For the clinical subset, all participants signed an informed-consent form that explicitly allows the use of their video and physiological data for non-commercial, academic research \emph{under access control}, but does not permit fully public, unrestricted distribution. This means the raw data cannot be posted on open file-sharing platforms without a data-use agreement.

To balance reproducibility and privacy, we will adopt an \emph{on-request} release model similar to widely used camera-based physiological datasets such as PURE~\cite{pure} and MMPD~\cite{tang2023mmpd}. Authorized researchers from recognized academic or medical institutions can request access by signing a joint usage and non-redistribution agreement that (1) restricts the data to non-commercial research, (2) prohibits re-identification or face recognition, and (3) forbids secondary sharing with third parties. This model has been successfully practiced in multiple international dataset projects and has been accepted by ethics boards in many institutions~\cite{li2018obf}.

We will release the raw facial and fingertip video files (i.e., without blurring) together with synchronized physiological CSV files so that rPPG algorithms can be trained and evaluated fairly on the original signals. At the same time, we \textbf{require} downstream users to apply minimal privacy protection when presenting qualitative examples in papers, talks, or online materials. In particular, for patient recordings, the eye region (or the whole face when necessary) should be blurred or masked in figures to prevent casual identification. This policy preserves the scientific utility of the dataset while honoring the restrictions in the patients’ consent forms.

\subsection{Motivation: Dual-View Mobile Sensing for Cardiovascular Patients}
Handheld smartphone videos suffer from view-specific failure modes: facial recordings are easily corrupted by head motion or suboptimal illumination, while fingertip recordings require stable contact that many elderly or cardiovascular patients cannot consistently maintain. These failures, however, are often \emph{complementary}—when the face view degrades, the fingertip may still retain a clean pulsatile signal, and vice versa. This observation motivates recording \emph{simultaneous} front- and rear-camera videos so that algorithms can dynamically rely on the better view instead of a single, potentially unreliable source.

At the same time, our clinical target population is physiologically more variable than the young, cooperative subjects commonly used in prior smartphone rPPG datasets. As summarized in Table~\ref{tab:data_stats}, the Lab subset provides long, well-instrumented dual-view sequences (over 360k facial and fingertip frames each) suitable for method development, whereas the Clinic subset contributes a large number of patient recordings with synchronized videos and medical references collected under real outpatient conditions. The demographic distribution in Fig.~\ref{fig:subject_distribution} further shows that the Lab participants are mostly young adults, while the Clinic cohort is dominated by older patients with a much wider range of height and weight, reflecting the actual diversity of cardiovascular users. This shift in age and body habitus is clinically relevant because skin properties, peripheral perfusion, and motion control all tend to deteriorate with age.

To quantify this clinical difficulty, we also compared heart-rate–variability (HRV) descriptors between the two subsets (Table~\ref{tab:hrv_comparison}). Patients in the Clinic subset exhibit significantly higher SDNN, SDSD, and SD2, indicating more irregular and less autonomically regulated rhythms than healthy volunteers. In such cases, relying on a \emph{single} peripheral site increases the risk of under-counting beats (pulse deficit). By capturing two vascular beds simultaneously—the facial region via the front camera and the fingertip via the rear camera with flash—M$^3$PD provides exactly the data needed to design fusion models that remain reliable when one view loses pulsatility. This is why we release M$^3$PD as a \emph{dual-view} and \emph{patient-inclusive} mobile dataset, rather than only a clean laboratory collection.

\begin{table}[h]
\centering
\caption{Comparison of HRV metrics between healthy subjects and cardiovascular patients. Values are Mean $\pm$ SD.}
\label{tab:hrv_comparison}
\begin{tabular}{lccc p{3.6cm}}
\toprule
\textbf{Indicator} & \textbf{Lab} & \textbf{Clinic} & \textbf{p-value} & \textbf{Note} \\
\midrule
SDNN & 34.823 ± 8.269 & 41.393 ± 7.360 & 0.013 
& Overall HRV: higher here mainly reflects rhythm irregularity in patients. \\
SDSD & 23.348 ± 6.345 & 26.687 ± 3.133 & 0.015 
& Short-term/beat-to-beat variability: more unstable cycles in clinic group. \\
SD2 & 23.401 ± 5.677 & 32.286 ± 6.177 & $<0.001$
& Long-term variability: suggests slower, irregular modulation. \\
PPA & 2698.219 ± 996.072 & 4174.720 ± 1276.353 & $<0.001$
& Poincaré area: more scattered RR pattern, harder for camera PPG. \\
\bottomrule
\end{tabular}
\end{table}

\section{Methodology}
In this section, we first describe the data preprocessing steps applied to both facial and fingertip videos (Section \ref{sec:preprocessing}). We then provide a detailed explanation of the network architecture (Section \ref{sec:overview}). Next, we discuss the view-specific branch (Section \ref{sec:modal_branch}) and the Fusion Mamba block (Section \ref{sec:fusion_mamba}) in depth. Finally, we describe the loss function (Section \ref{sec:loss_function}).

\subsection{Data Processing}
\label{sec:preprocessing}
To prepare the dual-view videos for physiological signal extraction, we applied several preprocessing steps to standardize the input data. Both facial and fingertip videos were resized to 128 × 128 pixels to maintain computational efficiency while preserving sufficient spatial information for capturing subtle color variations related to blood volume changes. For facial videos, we additionally applied face detection and cropping to extract stable facial regions, reducing the impact of background clutter and head movements during handheld recording. The temporal length of the input video segments was set to 160 frames, corresponding to approximately 5.3 seconds at 30 fps, which provides sufficient temporal context for capturing multiple cardiac cycles while maintaining computational tractability.

In the post-processing stage, we applied a bandpass filter (0.5-3 Hz, corresponding to 30-180 BPM) to the predicted PPG waveform to remove high-frequency noise and low-frequency baseline drift. Heart rate was then estimated from the filtered signal using the Welch method for power spectral density estimation~\cite{liu2023rppg}, identifying the dominant frequency component within the physiological range.

\subsection{Facial-Fingertip Fusion Mamba ($\mathrm{F}^3 \text{Mamba}$) Framework}
\label{sec:overview}


\begin{figure*}[t] 
  \centering
  \includegraphics[width=\textwidth]{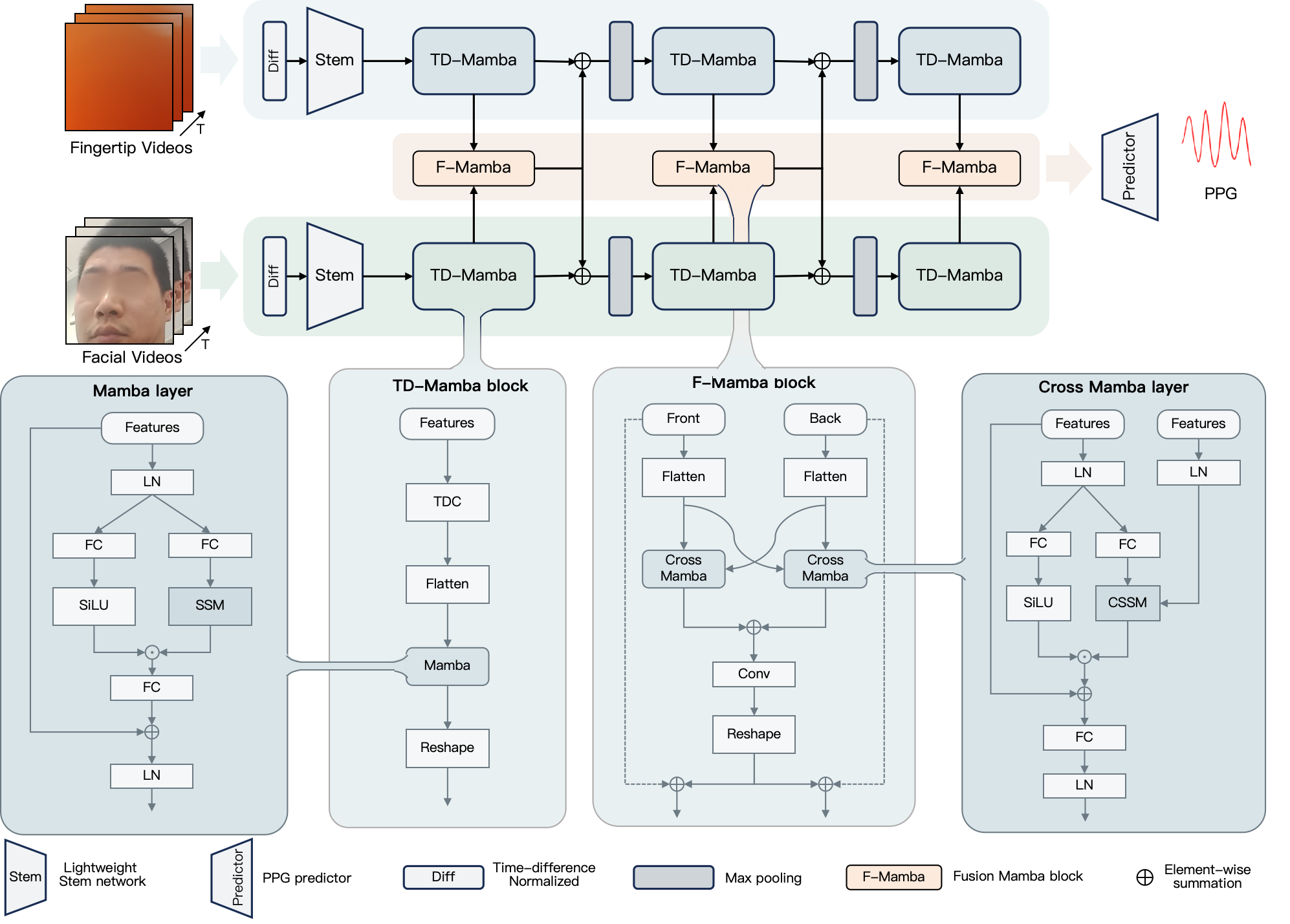}
  \caption{\textbf{The proposed $\mathrm{F}^3 \text{Mamba}$ framework.} The architecture processes facial and fingertip videos from front and rear cameras. After differentiation-normalization preprocessing, inputs flow through view-specific branches with TD-Mamba blocks for temporal modeling of each view separately. F-Mamba blocks facilitate cross-view fusion by integrating complementary information and dynamically updating state parameters. This fusion approach maintains robustness when one view contains artifacts, ultimately outputting a PPG waveform for heart rate estimation.}
  \label{fig:framework} 
\end{figure*}

To effectively integrate physiological signals from facial videos and fingertip videos for reliable cardiovascular monitoring, we propose a novel framework named $\mathrm{F}^3 \text{Mamba}$, as illustrated in \autoref{fig:framework}. This framework addresses common clinical challenges in mobile health monitoring by combining two complementary vascular beds - the facial region with rich microvasculature and the fingertip with dense capillary networks. Our design consists of parallel view-specific branches, a view fusion branch, and an rPPG predictor head. The view-specific branches process each input source using three Temporal Difference Mamba (TD-Mamba) blocks \cite{luo2024physmamba}, while the view fusion branch uses Fusion Mamba (F-Mamba) blocks to combine their physiological information. This approach maintains measurement continuity when one signal source is temporarily compromised by patient movement.

For processing, stable facial regions are first extracted from facial videos through cropping. These facial frames and the raw finger frames are then fed into the $\mathrm{F}^3 \text{Mamba}$ framework, where the DiffNormalized technique \cite{chen2018deepphys} extracts frame difference features to highlight subtle blood volume changes. These processed frames are sent to separate view-specific branches where a basic stem network extracts initial spatial features. Three TD-Mamba blocks then model the temporal patterns essential for capturing cardiovascular pulsations across video frames.

The multi-stage fusion branch is a key innovation that enables robust physiological monitoring in challenging daily healthcare clinical scenarios. This branch contains three F-Mamba blocks that align with the view-specific branches. Each F-Mamba block takes both facial and finger feature maps as input and creates a fused representation that captures the most reliable physiological signals from each source. These fused features connect back to their respective view branches through residual connections, allowing mutual enhancement of signal quality. After the final F-Mamba block, the combined features are processed by the rPPG predictor to generate the pulse waveform needed for heart rate calculation.

\subsection{Dual View (Front and Rear) Feature Extraction}
\label{sec:modal_branch}
The preprocessed video frames $F_{\text{pre}}^d$ are processed through a stem network consisting of three simple convolutional layers to achieve spatial downsampling:

\begin{equation}
F_{\text{stem}}^d = \text{stem}(F_{\text{pre}}^d)
\end{equation}

Where $F_{\text{stem}}^d \in \mathbb{R}^{C \times T \times H \times W}$ are the downsampled feature maps, $C$ denotes the channel dimension, $T$ is the temporal length, and $H/W$ represent the downsampled height and width, respectively. The subscript $d$ indicates the input view (face or fingertip). $F_{\text{stem}}^d$ is then fed into a view-specific branch for further feature extraction. 

Each branch consists of three TD-Mamba blocks, where each block integrates a temporal difference convolution (TDC) layer followed by a Mamba layer. The following provides a detailed explanation of these layers.
The TDC block efficiently captures fine-grained, local, spatio-temporal dynamics, which are crucial for tracking subtle color changes \cite{yu2021searching}:

\begin{equation}
\begin{split}
TDC(F_i^d) = \underbrace{\sum_{p_n \in \mathcal{C}} w(p_n) \cdot F_i^d(p_0 + p_n)}_{\text{vanilla 3D convolution}} + \theta \cdot \underbrace{\left( -F_i^d(p_0) \cdot \sum_{p_n \in \mathcal{R}''} w(p_n) \right)}_{\text{temporal CD term}}
\end{split}
\end{equation}

Where, $i\in\{0, 1, 2\}$ denotes the feature refinement stage and $F_i^d$ denotes the input feature of the TD-mamba block in stage $i$. $w$ is a learnable parameter, $p_0$, $C$ and $R''$ represent the current spatio-temporal location, the sampled local neighborhood, and the sampled adjacent neighborhood, respectively. Unlike vanilla convolution, TDC explicitly models temporal correction, enhancing its ability to extract time-dependent features. The TDC-processed features $F_i^d \in \mathbb{R}^{C \times T \times H_i \times W_i}$ are flattened to $F_i^d \in \mathbb{R}^{C \times (T H_i W_i)}$ and passed through a Mamba layer:

\begin{equation}
    x_i^d, z_i^d = \mathrm{Linear}_x\left(\mathrm{LN}\left(F_i^d\right)\right), \mathrm{Linear}_z\left(\mathrm{LN}\left(F_i^d\right)\right)
    \label{eq:linear_transformation}
\end{equation}

Here, LN denotes layer normalization, ${\rm Linear}_x$ and ${\rm Linear}_z$ represent two distinct fully connected layers. The transformed sequence $x_i^d$ is then processed through a Selective State Space (SSM) block for feature extraction, yielding the output sequence:

\begin{equation}
    y_i^d=SSM\left(x_i^d\right)
\end{equation}

After gating with ${z}_{i}^{d}$, the sequence is processed through a fully connected layer and added to the original $F_i^d$. We then use layer normalization on the output, resulting in the final output of Mamba layer:

\begin{equation}
    F_i^{Out,d} = \text{LN}(\text{Linear}_{out}\left(y_i^d \cdot \text{SiLU}\left(z_i^d\right)\right) + F_i^d
    \label{eq:example_label})
\end{equation}

Here, $\text{Linear}_{\text{out}}$ denotes the fully connected layer, and SiLU is the activation function. The output feature map $F_i^{\text{Out,d}} \in \mathbb{R}^{C \times (T H_i W_i)}$ from the Mamba layer is reshaped to $\mathbb{R}^{C \times T \times H_i \times W_i}$. Following each TD-Mamba block, a $2 \times 2$ max-pooling layer performs spatial downsampling, after three such TD-Mamba blocks, the final output feature map from each view branch is $F_3^{\text{Out,d}} \in \mathbb{R}^{C \times T \times \frac{H}{8} \times \frac{W}{8}}$.

\subsection{Aligned Facial and Fingertip Feature Fusion}
\label{sec:fusion_mamba}

To combine the dual-view features $F_i^{Out,d}$ extracted from TD-Mamba block, we introduce the F-Mamba block. This block processes paired facial and fingertip features as inputs and generates the fused representations $F_i^f \in \mathbb{R}^{C \times T \times H_i \times W_i}$. 


The F-Mamba block employs a parallel, dual-branch architecture to seamlessly integrate multi-view features. Each branch processes view-specific features $F_i^d \in \mathbb{R}^{C \times T \times H_i \times W_i}$, which are flattened along the temporal dimensions to form a new representation $\widetilde{F}_i^d \in \mathbb{R}^{C \times (TH_iW_i)}$. Cross-view interaction is facilitated through a symmetric cross Mamba layer, designed to operate on these flattened sequence representations. Unlike the standard Mamba layer, the cross Mamba layer incorporates a Cross State Space Model (CSSM), enabling bidirectional information exchange between complementary views. This mechanism dynamically updates each branch's state space, ensuring robust feature fusion, as detailed in Algorithm \autoref{tab:alg1}.

\begin{algorithm}
\caption{CSSM Block}  
\label{tab:alg1}
\begin{algorithmic}[1]
\Require $x^a, x^b$: $(B,L,D)$ \Comment{Input features}
\Ensure $y^a$: $(B,L,D)$ \Comment{Fused features}

\State $A: (B,D,N) \gets \text{Parameter}_A$ \Comment{Learnable state matrix}
\State $B_a: (B,L,N) \gets \text{Linear}_B(x^a)$ 
\State $B_b: (B,D,N) \gets \text{Linear}_B(x^b)$ 
\State $B: (B,L,N) \gets (1-\lambda_b)B_a + \lambda_b B_b$ \Comment{Fuse views}
\State $C_a: (B,L,N) \gets \text{Linear}_C(x^a)$ 
\State $C_b: (B,L,N) \gets \text{Linear}_C(x^b)$ 
\State $C: (B,L,N) \gets (1-\lambda_c)C_a + \lambda_c C_b$
\Comment{Fuse views}
\State $\Delta: (B,L,D) \gets \log(1+\exp(\text{Linear}_{\Delta}(x^a) + \text{Parameter}_{\Delta})$ \Comment{Softplus discretization}
\State ${\overline A}: (B,L,D,N) \gets \exp(\Delta \otimes A)$ \Comment{Discretize state transition}
\State ${\overline B}: (B,L,D,N) \gets \Delta \otimes B_a$ \Comment{Discretize input matrix}
\State $y^a \gets \text{SSM}({\overline A},{\overline B},C)(x^a)$ \Comment{State space computation}
\State \Return $y^a$ 
\end{algorithmic}
\end{algorithm}

\textbf{Variable Explanations:}
\begin{multicols}{2}
\begin{itemize}
\item $x^a$: Primary view input features
\item $x^b$: Complementary view input features
\item $y^a$: Enhanced output features of the primary view
\item $A$: Learnable state transition matrix
\item $B_a$: Input projection of primary view $x^a$
\item $B_b$: Input projection of complementary view $x^b$
\item $C$: Output projection matrix derived from $x^a$
\item $\Delta$: Discretization step size
\item $\overline{A}$: Discretized state transition matrix
\item $\overline{B}$: Discretized input matrix
\item $\lambda_b$: Fusion weight for input projection
\item $\lambda_c$: Fusion weight for output projection
\end{itemize}
\end{multicols}

The fusion feature \(F_i^f \in \mathbb{R}^{C \times (TH_iW_i)}\) is generated by the convolutional integration of these cross-view features:

\begin{equation}
  \widetilde F_i^f = \text{Conv}\left( \left[ \widetilde{F}_i^{d_1 \rightarrow d_2}+\widetilde{F}_i^{d_2 \rightarrow d_1} \right] \right)
\end{equation}

Here, $\widetilde{F}_i^{d_1 \rightarrow d_2}$ and $\widetilde{F}_i^{d_2 \rightarrow d_1}$ are outputs from the Cross Mamba layer. Each represents features where one view's state space ($d_1$ or $d_2$) is updated with information from the other view.

To maintain dimensional consistency, we reshape $\widetilde F_i^f$ to match $F_i^d$. The reshaped features are then added to the original features through residual connections:

\begin{equation}
  F_i^{d}=  F_i^d + \text{Reshape}\left( \widetilde F_i^f \right)
\end{equation}

With $i \in {0,1}$ denoting the fusion stages. In the final stage, the F-Mamba block output $F_3^f$ is processed through adaptive average pooling to reduce dimensions, followed by convolution to generate PPG predictions:

\begin{equation}
  {\text{PPG}} = \text{Conv}(\text{AdaptiveAvgPool}(F_3^f))
  \end{equation}

This dual-view approach combines advantages from both input sources. When facial videos contain motion artifacts or poor lighting, fingertip data with its dense capillary network provides more stable signals. When fingertip contact is unstable, facial vasculature data maintains measurement quality. This complementary design improves heart rate estimation across various real-world conditions.

\subsection{Time-frequency Combined Loss Function}
\label{sec:loss_function}

To ensure robust learning, the loss function combines time-domain loss and frequency-domain loss, inspired by prior work \cite{zou2025rhythmformer}. Specifically, we employ the negative Pearson (NP) loss as the time-domain loss, defined as:

\begin{equation}
  \mathcal{L}_{{time}} = -\frac{\sum_{i=1}^{N} (y_i - \bar{y})(\hat{y}_i - \bar{\hat{y}})}{\sqrt{\sum_{i=1}^{N} (y_i - \bar{y})^2} \sqrt{\sum_{i=1}^{N} (\hat{y}_i - \bar{\hat{y}})^2}},
\end{equation}

where $y_i$ and $\hat{y}_i$ represent the ground truth and predicted signals, respectively, $\bar{y}$ and $\bar{\hat{y}}$ are their means, and $N$ is the number of samples.

The frequency-domain loss $\mathcal{L}_{{freq}}$ is calculated using the cross-entropy (CE) between the aligned power spectrum of the predicted signal and the ground truth signal. To align the power spectra, the index of the maximum value in the ground truth spectrum is used as a reference. It is expressed as:

\begin{equation}
  \mathcal{L}_{{freq}} = \text{CE}(\text{MaxIndex}(y_{PSD}, \hat{y}_{PSD}))
\end{equation}

where $\text{MaxIndex}$ indicates the index of the maximum value, and $y_{PSD}$, $\hat{y}_{PSD}$ represent the power spectrum decomposition (PSD) of the ground true signals $y$ and the predicted signals $\hat{y}$. Additionally, we introduce a PSD distribution loss $\mathcal{L}_{PSD}$ to constrain the predicted PSD distribution to closely match the ground truth PSD distribution. This loss is defined using the Kullback-Leibler (KL) divergence as:

\begin{equation}
  \mathcal{L}_{PSD} = \text{KL}\left(P(y_{PSD}) \| P(\hat{y}_{PSD})\right)
\end{equation}

In this equation, $P(y_{PSD})$ and $P(\hat{y}_{PSD})$ represent the probability distributions of the ground truth and predicted PSD values, respectively. The KL divergence measures the difference between these two distributions, ensuring that the predicted HR aligns closely with the ground truth. Finally, the overall loss function is formulated as a weighted sum of the three components:

\begin{equation}
    \mathcal{L}_{\text{total}} = \lambda_1 \mathcal{L}_{\text{time}} + \lambda_2 \mathcal{L}_{\text{freq}} + \lambda_3 \mathcal{L}_{\text{PSD}}
\end{equation}

Where $\lambda_1$, $\lambda_2$, and $\lambda_3$ are hyperparameters that control the relative importance of each loss component. Following the parameter selection strategy in prior work~\cite{zou2025rhythmformer}, we set $\lambda_1 = 0.2$, $\lambda_2 = 1$, and $\lambda_3 = 1$ in our experiments.

\section{Results}

\subsection{Experimental Settings and Evaluation Metrics}
The programs were developed in Python 3.8 using PyTorch and executed on NVIDIA GeForce RTX 3090 GPUs. The TD-Mamba block followed the original settings from \cite{luo2024physmamba}. In the Fusion Mamba block, the CSSM dimension was set to 64 with a dropout rate of 0.1, the hyperparameters $\lambda_b$ and $\lambda_c$ were set to 0.5, indicating equal importance for both views in the fusion process.

We trained the model for 15 epochs with a batch size of 4, using the Adam optimizer with an initial learning rate of $5{\times}10^{-5}$ and employing a OneCycleLR scheduler to adjust the learning rate dynamically. The subjects were randomly divided into three groups, and a three-fold cross-validation protocol was performed to obtain the average results.

To assess the accuracy of HR measurement, we used the mean absolute error (MAE), mean absolute percentage error (MAPE), root mean squared error (RMSE), and Pearson correlation coefficient ($\rho$) as the evaluation metrics. All reported values are rounded to three decimal places for readability. Best results are presented in 
\textbf{bold}, and sub-optimal results are \underline{underlined}.

\subsection{Evaluation of Single-View Video Physiology with Existing Methods}
To establish a clear baseline for our dual-view fusion approach, we first analyze the performance of traditional unsupervised rPPG or contact PPG (cPPG) methods on single-view inputs. These methods, including GREEN \cite{verkruysse2008remote}, ICA \cite{poh2010advancements}, POS \cite{wang2016algorithmic}, and OMIT \cite{casado2023face2ppg}, serve as common baselines to highlight the inherent limitations of relying on a single camera view in real-world handheld scenarios.

While traditional methods provide a foundational baseline, recent advancements in deep learning have led to state-of-the-art (SOTA) performance in rPPG/cPPG. We evaluate several deep learning models, including PhysNet~\cite{yu2019remote}, PhysFormer~\cite{yu2022physformer}, RhythmFormer~\cite{zou2025rhythmformer}, and PhysMamba~\cite{luo2024physmamba}. For a fair comparison, all deep learning models are trained on the PURE dataset \cite{stricker2014non}. To be noted, we evaluate fingertip single-view tasks using the deep learning models trained on PURE since there are few accessible fingertip video datasets with pre-trained models.

\subsubsection{Challenges with Face rPPG in Handheld Scenarios}
As shown in \autoref{tab:front_view_traditional} and \autoref{tab:front_view_deeplearning}, traditional methods perform poorly on facial videos in our handheld setting, with MAEs ranging from 16.5 to 32.9 BPM and near-zero correlation ($\rho$), reflecting high sensitivity to motion and illumination. Deep learning baselines trained on PURE also show a marked degradation on $\mathrm{M}^3\text{PD}$; on the Lab dataset, the best model achieves an MAE of 15.0 BPM, while on the Clinic dataset, MAEs are between 23 and 29 BPM, evidencing a strong domain shift.

\begin{figure*}[htbp]
  \centering
  \begin{subfigure}{\linewidth}
    \centering
    \includegraphics[width=\linewidth]{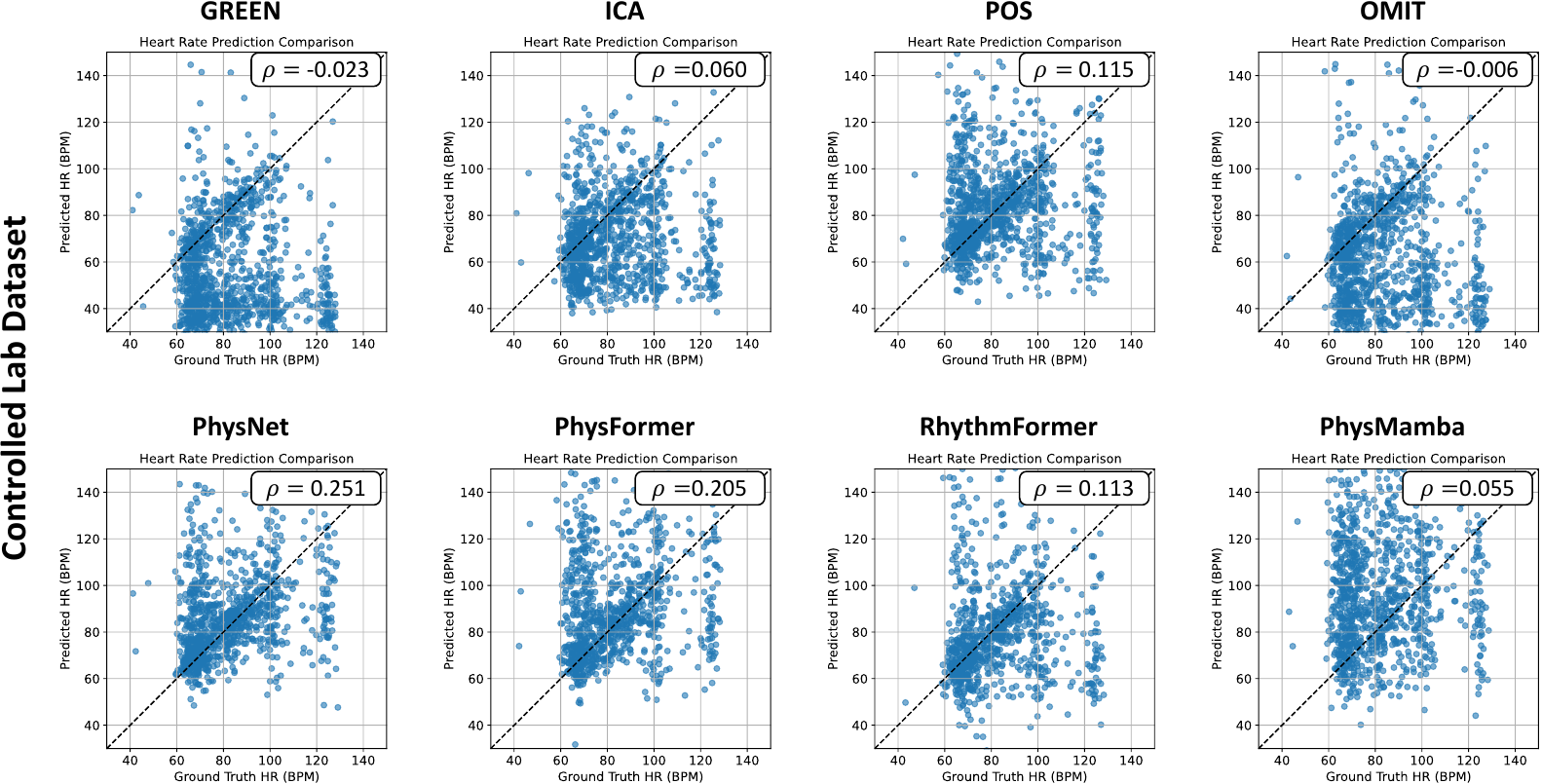}
    \caption{Lab Dataset (Face inputs).}
    \label{subfig:front_view_lab}
  \end{subfigure}
  \vspace{0.5em}
  \begin{subfigure}{\linewidth}
    \centering
    \includegraphics[width=\linewidth]{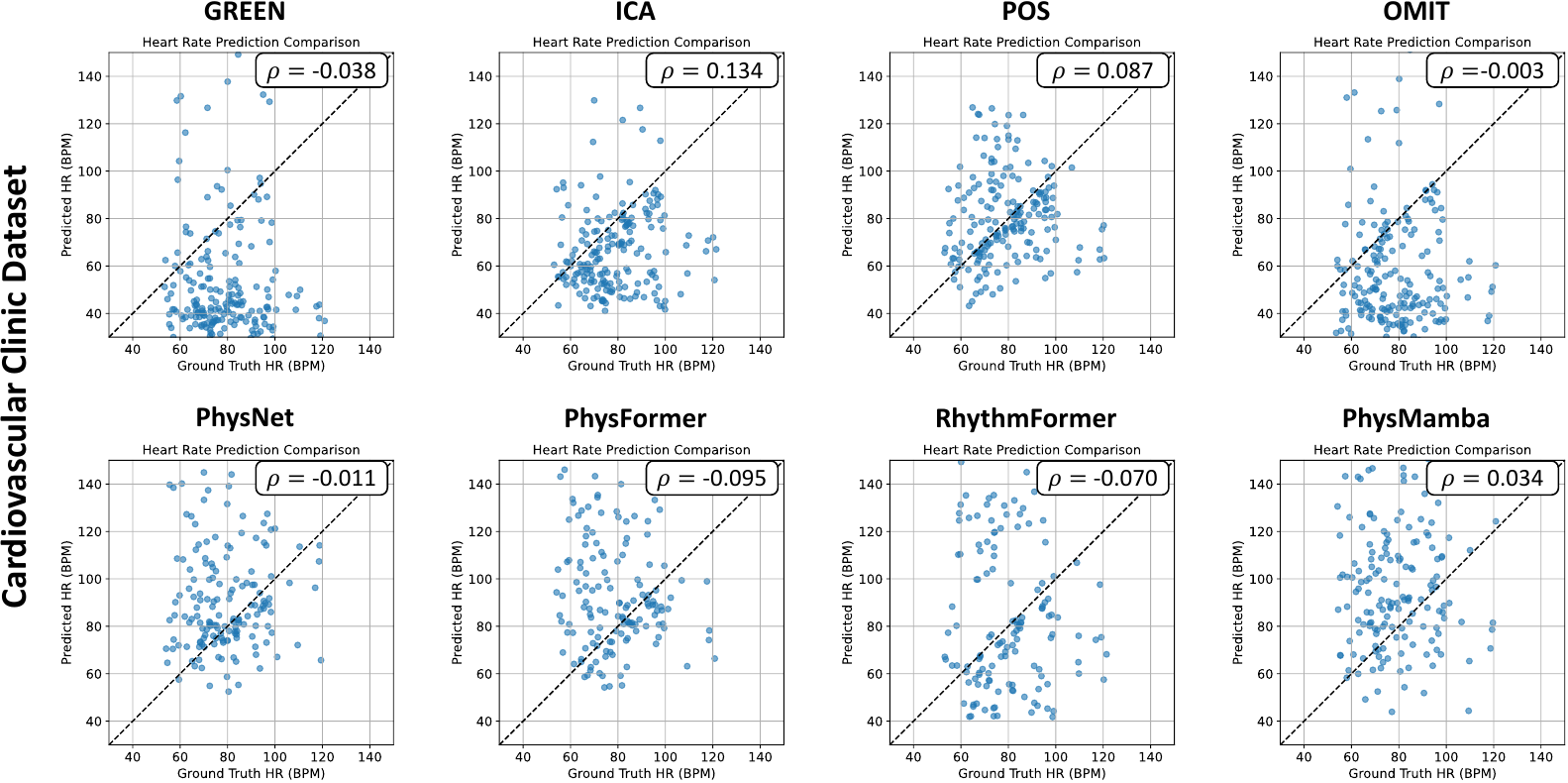}
    \caption{Clinic Dataset (Face inputs).}
    \label{subfig:front_view_clinic}
  \end{subfigure}
  \caption{\textbf{Scatter plots of estimated HR vs. reference HR for face inputs.} Top: Lab; Bottom: Clinic. The diagonal indicates ideal correlation.}
  \label{fig:front_view_scatter}
\end{figure*}

To further illustrate these limitations, we visualize the heart rate estimation results using scatter plots in \autoref{fig:front_view_scatter}. Ideally, the estimated heart rate should be highly correlated with the reference heart rate, with data points closely distributed along the diagonal. However, the scatter plots for both datasets reveal significant deviation and dispersion, indicating poor correlation and large estimation errors. This demonstrates that single-view facial rPPG under handheld motion and lighting variations is unstable for accurate heart rate measurement.

\begin{table*}[htbp]
  \centering
  \caption{Performance of Traditional Unsupervised Methods on the $\mathrm{M}^3\text{PD}$ Dataset Using Face Inputs.}\label{tab:front_view_traditional}
  \resizebox{\textwidth}{!}{%
  \begin{tabular}{ccccccccc}  
      \toprule[1.2pt]
      \multirow{2}{*}{\textbf{Method}} & \multicolumn{4}{c}{\textbf{Lab}} & \multicolumn{4}{c}{\textbf{Clinic}} \\
      \cmidrule(lr{3pt}){2-5}  
      \cmidrule(l{3pt}r){6-9}  
      & \textbf{MAE} $\downarrow$ & \textbf{MAPE} $\downarrow$ & \textbf{RMSE} $\downarrow$ & $\rho$ $\uparrow$ & \textbf{MAE} $\downarrow$ & \textbf{MAPE} $\downarrow$ & \textbf{RMSE} $\downarrow$ & $\rho$ $\uparrow$ \\
      \midrule[0.8pt]
        GREEN & 28.924 & 32.966 & 37.173 & -0.023 & 32.892 & 40.550 & 38.202 & -0.038 \\
        ICA & \underline{19.438} & \underline{21.902} & \underline{26.897} & \underline{0.060} & \underline{18.531} & \underline{22.627} & \underline{24.079} & \textbf{0.134} \\
        POS & \textbf{16.549} & \textbf{19.851} & \textbf{23.755} & \textbf{0.115} & \textbf{16.475} & \textbf{21.190} & \textbf{22.210} & \underline{0.087} \\
        OMIT & 25.672 & 29.145 & 34.364 & -0.006 & 28.189 & 34.704 & 34.473 & -0.003 \\   \bottomrule[1.2pt]
  \end{tabular}}
  \\ \vspace{0.1cm}
\small
MAE = Mean Absolute Error in HR estimation (Beats/Min), RMSE = Root Mean Square Error in HR estimation (Beats/Min), MAPE = Mean Percentage Error (\%), $\rho$ = Pearson Correlation in HR estimation.
\end{table*}

\begin{table*}[htbp]
  \centering
  \caption{Performance of Deep Learning Methods on the $\mathrm{M}^3\text{PD}$ Dataset Using Face Inputs.}\label{tab:front_view_deeplearning}
  \resizebox{\textwidth}{!}{%
  \begin{tabular}{ccccccccc}  
      \toprule[1.2pt]
      \multirow{2}{*}{\textbf{Method}} & \multicolumn{4}{c}{\textbf{PURE$\to$Lab}} & \multicolumn{4}{c}{\textbf{PURE$\to$Clinic}} \\
      \cmidrule(lr{3pt}){2-5}  
      \cmidrule(l{3pt}r){6-9}  
      & \textbf{MAE} $\downarrow$ & \textbf{MAPE} $\downarrow$ & \textbf{RMSE} $\downarrow$ & $\rho$ $\uparrow$ & \textbf{MAE} $\downarrow$ & \textbf{MAPE} $\downarrow$ & \textbf{RMSE} $\downarrow$ & $\rho$ $\uparrow$ \\
      \midrule[0.8pt]
        PhysNet & \textbf{15.041} & \textbf{18.633} & \textbf{22.501} & \textbf{0.251} & \underline{23.497} & \underline{33.361} & \textbf{32.426} & \underline{-0.011} \\
        PhysFormer & \underline{17.186} & 21.457 & \underline{24.652} & \underline{0.205} & \textbf{23.473} & \textbf{31.296} & \underline{32.450} & -0.095 \\
        RhythmFormer & 17.239 & \underline{20.184} & 25.974 & 0.113 & 29.337 & 38.062 & 38.230 & -0.070 \\
        PhysMamba & 24.998 & 31.988 & 32.101 & 0.055 & 26.590 & 33.682 & 35.622 & \textbf{0.034} \\   \bottomrule[1.2pt]
  \end{tabular}}
  \\ \vspace{0.1cm}
\small
MAE = Mean Absolute Error in HR estimation (Beats/Min), RMSE = Root Mean Square Error in HR estimation (Beats/Min), MAPE = Mean Percentage Error (\%), $\rho$ = Pearson Correlation in HR estimation.
\end{table*}

\subsubsection{Challenges with Fingertip cPPG in Handheld Scenarios}
The performance on fingertip inputs, detailed in~\autoref{tab:back_view_traditional} and~\autoref{tab:back_view_deeplearning}, reveals that while certain methods like ICA show improved accuracy on the Lab dataset compared to facial inputs (with an MAE of 7.749 BPM), the overall results remain inconsistent across different algorithms and datasets and fail to achieve the desired high precision. This variability underscores the limitations of relying on a single view, as factors like finger placement and pressure significantly impact signal quality, highlighting the reliability challenges that persist in practical handheld scenarios.

\begin{figure*}[htbp]
  \centering
  \begin{subfigure}{\linewidth}
    \centering
    \includegraphics[width=\linewidth]{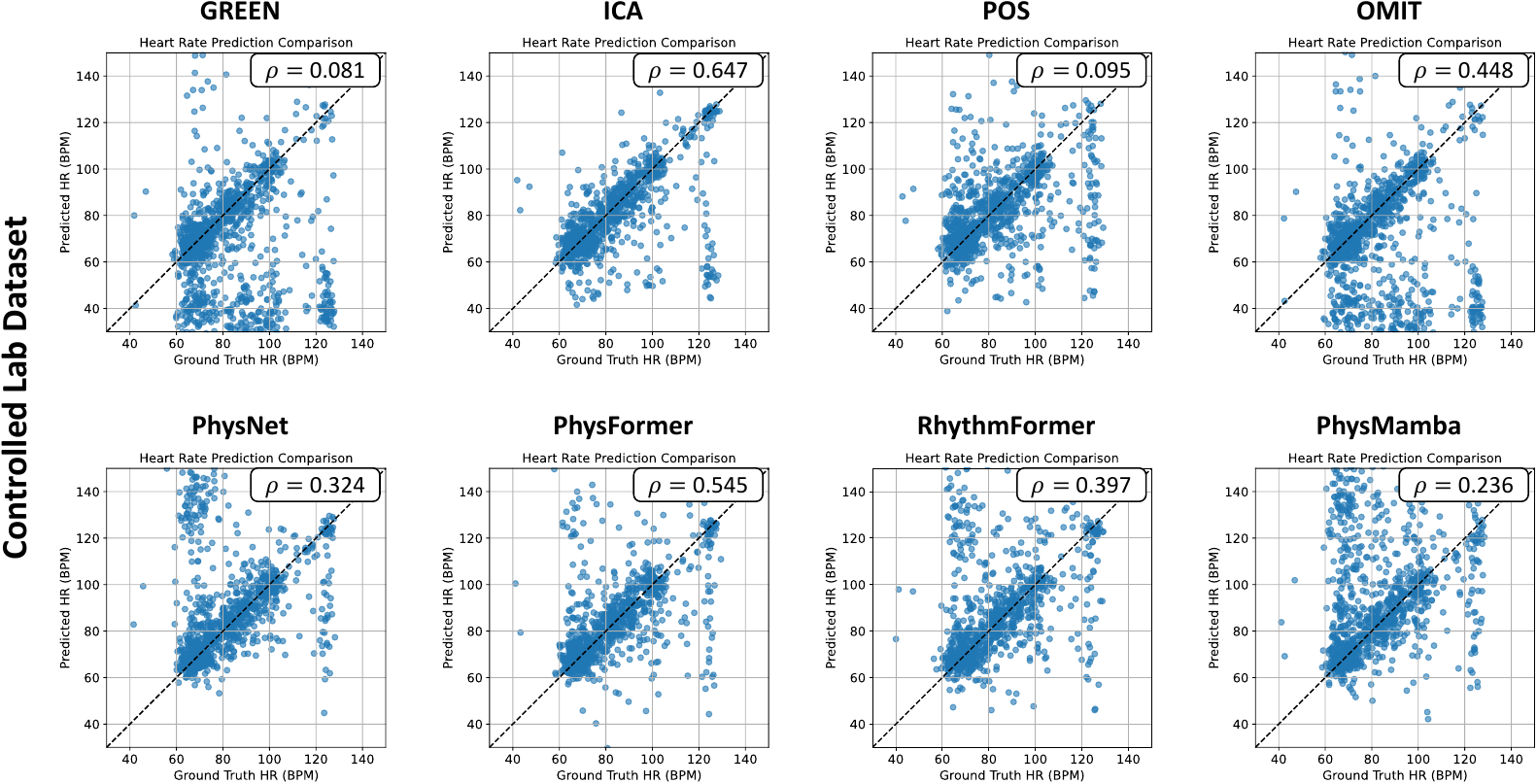}
    \caption{Lab Dataset (Fingertip inputs).}
    \label{subfig:back_view_lab}
  \end{subfigure}
  \vspace{0.5em}
  \begin{subfigure}{\linewidth}
    \centering
    \includegraphics[width=\linewidth]{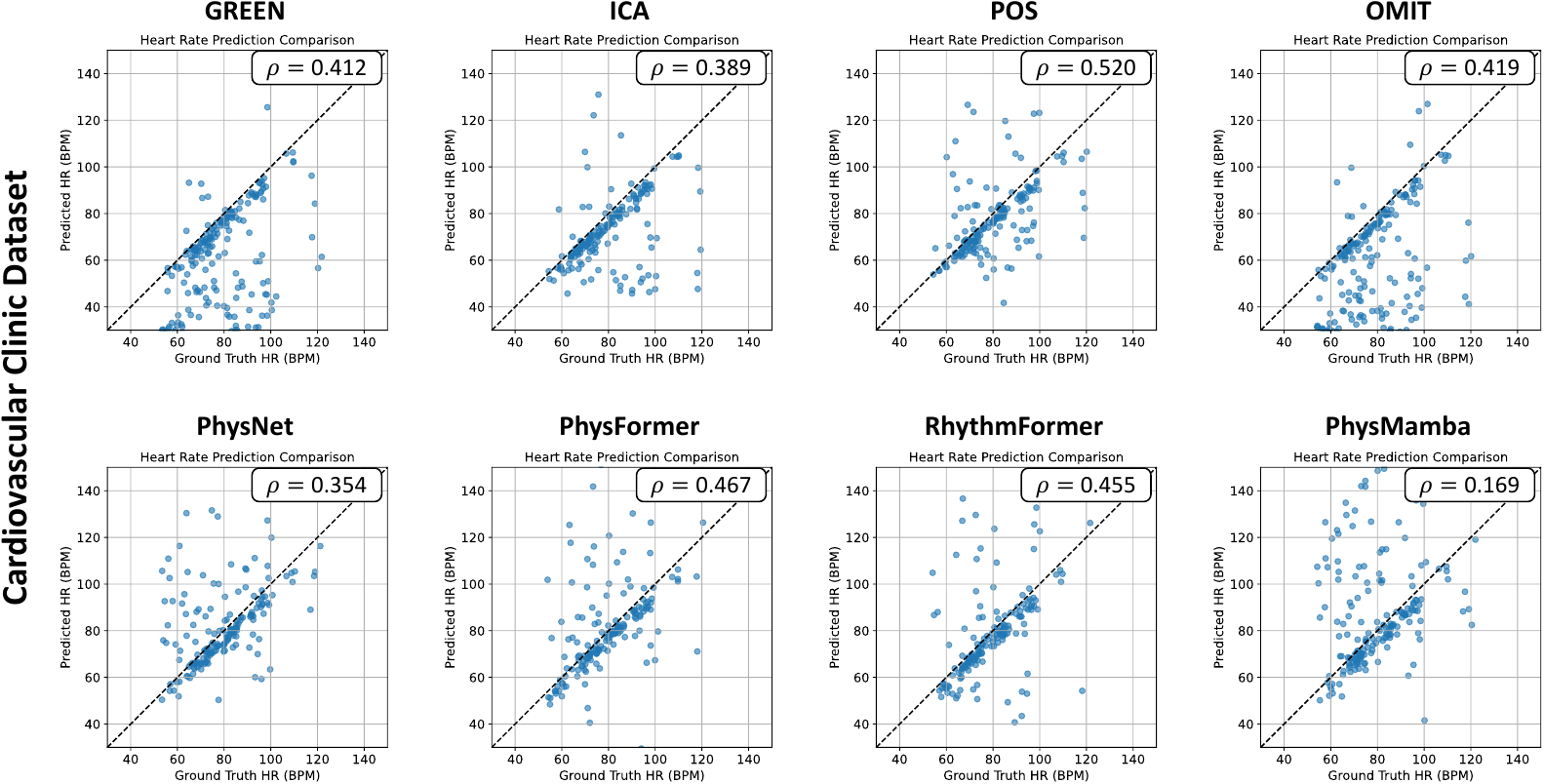}
    \caption{Clinic Dataset (Fingertip inputs).}
    \label{subfig:back_view_clinic}
  \end{subfigure}
  \caption{\textbf{Scatter plots of estimated HR vs. reference HR for fingertip inputs.} Top: Lab; Bottom: Clinic. The diagonal indicates ideal correlation.}
  \label{fig:back_view_scatter}
\end{figure*}

To further illustrate these limitations, we visualize the heart rate estimation results in \autoref{fig:back_view_scatter}. The scatter plots reveal considerable spread and deviation, indicating that even fingertip cPPG is susceptible to significant estimation errors and poor correlation under handheld conditions. This instability is often caused by incomplete finger coverage, variable pressure, or movement during measurement factors that are common in real-world use, especially among elderly patients. These results demonstrate that single-view fingertip cPPG, while sometimes more robust than facial rPPG, still faces significant reliability challenges in practical handheld scenarios.

\begin{table*}[htbp]
  \centering
  \caption{Performance of Traditional Unsupervised Methods on the $\mathrm{M}^3\text{PD}$ Dataset Using Fingertip Inputs.}\label{tab:back_view_traditional}
  \resizebox{\textwidth}{!}{%
  \begin{tabular}{ccccccccc}  
      \toprule[1.2pt]
      \multirow{2}{*}{\textbf{Method}} & \multicolumn{4}{c}{\textbf{Lab}} & \multicolumn{4}{c}{\textbf{Clinic}} \\
      \cmidrule(lr{3pt}){2-5}  
      \cmidrule(l{3pt}r){6-9}  
      & \textbf{MAE} $\downarrow$ & \textbf{MAPE} $\downarrow$ & \textbf{RMSE} $\downarrow$ & $\rho$ $\uparrow$ & \textbf{MAE} $\downarrow$ & \textbf{MAPE} $\downarrow$ & \textbf{RMSE} $\downarrow$ & $\rho$ $\uparrow$ \\
      \midrule[0.8pt]
        GREEN & 19.735 & 22.213 & 31.380 & 0.081 & 17.468 & 21.775 & 25.045 & \underline{0.421} \\
        ICA & \textbf{7.749} & \textbf{8.924} & \textbf{14.763} & \textbf{0.647} & \underline{10.881} & \underline{12.720} & \underline{18.126} & 0.389 \\
        POS & 18.461 & 20.659 & 30.452 & 0.095 & \textbf{9.423} & \textbf{11.565} & \textbf{14.845} & \textbf{0.520} \\
        OMIT & \underline{11.011} & \underline{12.835} & \underline{18.053} & \underline{0.448} & 18.511 & 23.168 & 26.219 & 0.419 \\   \bottomrule[1.2pt]
  \end{tabular}}
  \\ \vspace{0.1cm}
\small
MAE = Mean Absolute Error in HR estimation (Beats/Min), RMSE = Root Mean Square Error in HR estimation (Beats/Min), MAPE = Mean Percentage Error (\%), $\rho$ = Pearson Correlation in HR estimation.
\end{table*}

\begin{table*}[htbp]
  \centering
  \caption{Performance of Deep Learning Methods on the $\mathrm{M}^3\text{PD}$ Dataset Using Fingertip Inputs.}\label{tab:back_view_deeplearning}
  \resizebox{\textwidth}{!}{%
  \begin{tabular}{ccccccccc}  
      \toprule[1.2pt]
      \multirow{2}{*}{\textbf{Method}} & \multicolumn{4}{c}{\textbf{PURE$\to$Lab}} & \multicolumn{4}{c}{\textbf{PURE$\to$Clinic}} \\
      \cmidrule(lr{3pt}){2-5}  
      \cmidrule(l{3pt}r){6-9}  
      & \textbf{MAE} $\downarrow$ & \textbf{MAPE} $\downarrow$ & \textbf{RMSE} $\downarrow$ & $\rho$ $\uparrow$ & \textbf{MAE} $\downarrow$ & \textbf{MAPE} $\downarrow$ & \textbf{RMSE} $\downarrow$ & $\rho$ $\uparrow$ \\
      \midrule[0.8pt]
        PhysNet & 13.021 & 17.017 & 23.689 & 0.324 & \underline{11.312} & \underline{18.859} & \underline{15.581} & 0.354 \\
        PhysFormer & \textbf{9.637} & \textbf{11.761} & \textbf{17.333} & \textbf{0.545} & \textbf{10.532} & \textbf{17.661} & \textbf{13.694} & \textbf{0.467} \\
        RhythmFormer & \underline{12.966} & \underline{16.302} & \underline{22.880} & \underline{0.397} & 12.845 & 22.302 & 16.182 & \underline{0.455} \\
        PhysMamba & 17.749 & 22.907 & 28.425 & 0.236 & 16.715 & 26.248 & 22.784 & 0.169 \\   \bottomrule[1.2pt]
  \end{tabular}}
  \\ \vspace{0.1cm}
\small
MAE = Mean Absolute Error in HR estimation (Beats/Min), RMSE = Root Mean Square Error in HR estimation (Beats/Min), MAPE = Mean Percentage Error (\%), $\rho$ = Pearson Correlation in HR estimation.
\end{table*}

\subsection{$\mathrm{F}^3 \text{Mamba}$Performs best with Dual-View Fusion}
Having established the limitations of traditional single-view methods, we now evaluate the performance of modern deep learning approaches, culminating in our proposed dual-view fusion model, $\mathrm{F}^3 \text{Mamba}$. This section details the results of intra-dataset and cross-dataset testing, demonstrating the superior accuracy and robustness achieved by integrating complementary physiological signals from both facial and fingertip videos.

\subsubsection{Intra-Dataset Testing}

We conducted intra-dataset tests on the two subsets of $\mathrm{M}^3 \text{PD}$ dataset (\textbf{Lab} and \textbf{Clinic}) and compared our results with deep learning methods\cite{luo2024physmamba,yu2022physformer,zou2025rhythmformer,yu2019remote}. All method parameters followed the original settings. We employed a three-fold cross-subject validation to prevent data leakage.

As shown in \autoref{tab:intra}, our model consistently outperforms existing approaches across both datasets. In the Lab dataset, our dual-view fusion approach reduces MAE by 30.2\% (from 9.542 to 6.664 BPM) compared to the best single-camera baseline. In the Clinic dataset with cardiovascular patients, we observe 21.9\% MAE reduction (from 9.480 to 7.405 BPM). This level of accuracy is particularly important in clinical cardiovascular monitoring, where heart rate measurement precision directly impacts arrhythmia detection and treatment decisions.

\begin{table*}[htbp]
  \centering
  \caption{Intra-dataset testing results on subsets of $\mathrm{M}^3\text{PD}$}\label{tab:intra}
  \resizebox{\textwidth}{!}{%
  \begin{tabular}{cccccccccc}  
      \toprule[1.2pt]  
      \multirow{2}{*}{\textbf{Method}} & \multirow{2}{*}{\textbf{Input}} & \multicolumn{4}{c}{\textbf{Lab}} & \multicolumn{4}{c}{\textbf{Clinic}} \\
      \cmidrule(lr{3pt}){3-6}  
      \cmidrule(l{3pt}r){7-10}  
      && \textbf{MAE} $\downarrow$ & \textbf{MAPE} $\downarrow$ & \textbf{RMSE} $\downarrow$ & $\rho$ $\uparrow$ & \textbf{MAE} $\downarrow$ & \textbf{MAPE} $\downarrow$ & \textbf{RMSE} $\downarrow$ & $\rho$ $\uparrow$ \\
      \midrule[0.8pt]
      \multirow{2}{*}{PhysNet} 
        & Face  & 31.651 & 37.350 & 39.238 & -0.057 & 25.159 & 32.158 & 30.951 & 0.047 \\
        & Finger & 10.325 & 10.464 & 19.563 & \textbf{0.640} & 16.476 & 20.971 & 22.738 & 0.385 \\
      \multirow{2}{*}{PhysFormer} 
        & Face  & 23.691 & 27.268 & 28.923 & 0.031  & 19.570 & 26.432 & 23.933 & 0.094 \\
        & Finger & 16.054 & 17.242 & 24.834 & 0.363  & 13.885 & 17.384 & 17.447 & 0.350 \\
      \multirow{2}{*}{RhythmFormer} 
        & Face  & 26.633 & 30.341 & 34.772 & 0.014  & 28.157 & 37.103 & 34.190 & -0.241 \\
        & Finger & 21.790 & 23.571 & 29.379 & 0.025  & 24.107 & 31.836 & 31.081 & -0.341 \\
      \multirow{2}{*}{PhysMamba} 
        & Face  & 14.041 & 13.341 & 22.759 & 0.428  & 15.481 & 20.269 & 20.032 & 0.032 \\
        & Finger & \underline{9.542}  & \underline{9.247}  & \underline{18.088} & 0.630  & \underline{9.480}  & \underline{11.411} & \underline{15.524} & \underline{0.460} \\
      \midrule
      $\mathrm{F}^3 \text{Mamba}$ {\textbf{(Ours)}} & Face+Finger & \textbf{6.664} & \textbf{6.859} & \textbf{12.796} & \underline{0.636} & \textbf{7.405} & \textbf{9.308} & \textbf{10.669} & \textbf{0.753} \\
      \bottomrule[1.2pt]  
  \end{tabular}}
  \\ \vspace{0.1cm}
\small
MAE = Mean Absolute Error in HR estimation (Beats/Min), RMSE = Root Mean Square Error in HR estimation (Beats/Min), MAPE = Mean Percentage Error (\%), $\rho$ = Pearson Correlation in HR estimation.
\end{table*}

\begin{figure*}[t] 
  \centering
  \includegraphics[width=\textwidth]{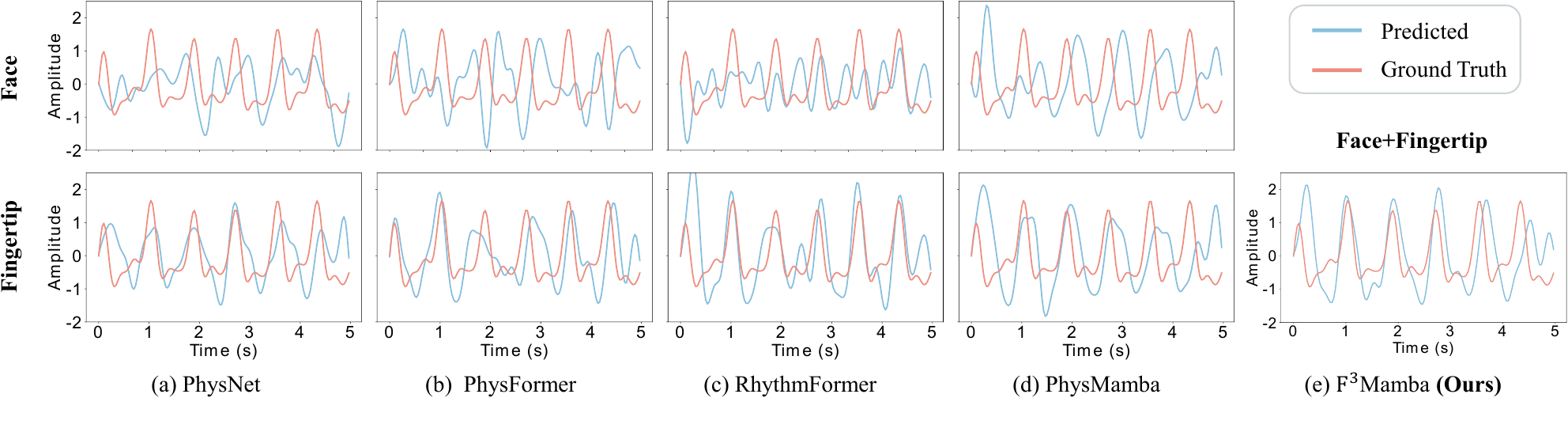}
  \caption{\textbf{Comparison between the predicted and ground-truth waveforms in the Lab dataset.} First row: Use facial view as input, second row: Use fingertip view as input. The predicted PPG waveforms generated by our $\mathrm{F}^3 \text{Mamba}$ framework show better consistency and accuracy.}
  \label{fig:visual ppg} 
\end{figure*}

\begin{figure}[tbp]
  \centering
  \begin{subfigure}{0.45\linewidth}
    \centering
    \includegraphics[width=\linewidth]{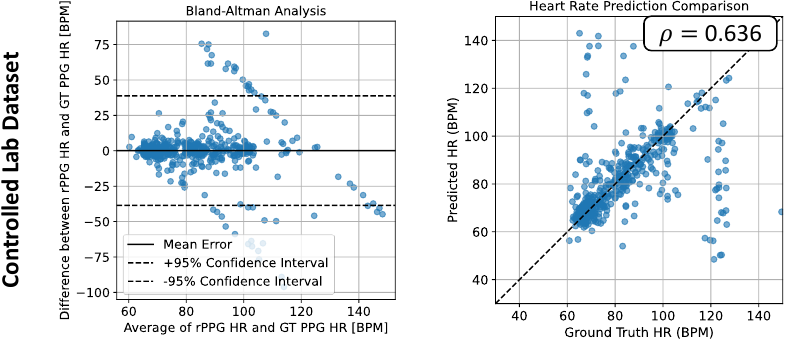}
    \caption{Lab Dataset}
    \label{subfig:ba_lab}
  \end{subfigure}
  \begin{subfigure}{0.45\linewidth}
    \centering
    \includegraphics[width=\linewidth]{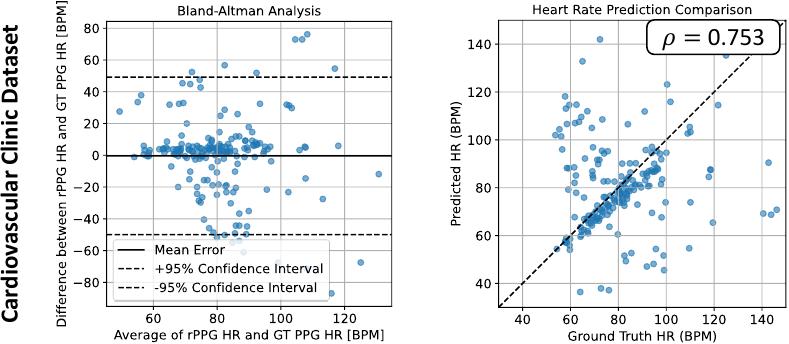}
    \caption{Clinic Dataset}
    \label{subfig:ba_clinic}
  \end{subfigure}

  \caption{\textbf{Bland-Altman analysis of $\mathrm{F}^3\text{Mamba}$ on intra-dataset $\mathrm{M}^3\text{PD}$}: (a) Lab dataset and (b) Clinic dataset. Mean error and 95\% limits of agreement are shown. Lighter colors indicate denser distributions; small random jitter is added to reduce overlap.}
  \label{fig:ba_hr_compare}
\end{figure}

We visualize the predicted and ground-truth waveforms in \autoref{fig:visual ppg}. The first row shows results using facial video input, while the second row shows fingertip video input results. The waveforms of our model closely match the ground truth, demonstrating accurate heart rate estimation through view fusion. Additionally, the Bland-Altman plot in \autoref{fig:ba_hr_compare} shows that 91.01\% of measurements fall within the 95\% confidence interval, with color intensity indicating data density. The balanced distribution around zero suggests no systematic bias, confirming measurement reliability for clinical applications.

Existing methods performed poorly on facial video inputs across both datasets due to two key clinical challenges. First, handheld smartphone recordings introduce motion artifacts that distort facial blood volume signals, a common issue in patient self-monitoring. Second, non-frontal recording angles (often from below) caused image distortion, mimicking real-world scenarios where patients struggle to maintain ideal device positioning.

When comparing input views, fingertip videos produced better results than facial videos across both datasets. This is consistent with physiological principles, as fingertips contain denser capillary networks with more direct hemodynamic signals and fewer confounding factors than facial regions. However, fingertip-based models still faced practical limitations in clinical use: patients often could not maintain steady contact with the rear camera, causing light interference and signal loss. Additionally, most algorithms were originally designed for facial video processing and could not fully capture the unique vascular properties of fingertip signals. Our fusion approach addresses these complementary limitations by maintaining measurement continuity when either source is temporarily compromised.

\subsubsection{Cross-Dataset Testing}

\begin{table*}[htbp]
  \centering
  \caption{Cross-dataset testing results on subsets of $\mathrm{M}^3\text{PD}$}\label{tab:cross}
  \resizebox{\textwidth}{!}{%
  \begin{tabular}{cccccccccc}  
      \toprule[1.2pt]
      \multirow{2}{*}{\textbf{Method}} & \multirow{2}{*}{\textbf{Input}} & \multicolumn{4}{c}{\textbf{Lab$\to$Clinic}} & \multicolumn{4}{c}{\textbf{Clinic$\to$Lab}} \\
      \cmidrule(lr{3pt}){3-6}  
      \cmidrule(l{3pt}r){7-10}  
      && \textbf{MAE} $\downarrow$ & \textbf{MAPE} $\downarrow$ & \textbf{RMSE} $\downarrow$ & $\rho$ $\uparrow$ & \textbf{MAE} $\downarrow$ & \textbf{MAPE} $\downarrow$ & \textbf{RMSE} $\downarrow$ & $\rho$ $\uparrow$ \\
      \midrule[0.8pt]
      \multirow{2}{*}{PhysNet} 
        & Face  & 21.177 & 29.079 & 28.409 & 0.322 & 27.234 & 34.283 & 34.287 & -0.143 \\
        & Finger & 24.383 & 31.654 & 38.786 & 0.102 & 18.537 & 21.579 & 27.728 & 0.143 \\
      \multirow{2}{*}{PhysFormer} 
        & Face  & 15.926 & 21.773 & 19.831 & 0.106 & 18.771 & 23.137 & 23.594 & 0.008 \\
        & Finger & 14.673 & 19.173 & 19.693 & 0.099 & 15.789 & 19.238 & 22.590 & 0.120 \\
      \multirow{2}{*}{RhythmFormer} 
        & Face  & 18.431 & 23.582 & 26.705 & 0.263 & 21.250 & 25.244 & 27.542 & -0.041 \\
        & Finger & 15.489 & 19.822 & 19.413 & -0.090 & 19.160 & 22.908 & 25.616 & -0.085 \\
      \multirow{2}{*}{PhysMamba} 
        & Face  & 12.352 & 16.917 & 16.776 & 0.274 & 14.053 & 16.740 & 19.352 & 0.218 \\
        & Finger & \underline{8.629} & \underline{10.840} & \underline{12.850} & \underline{0.599} & \textbf{8.522} & \textbf{9.302} & \underline{15.640} & \underline{0.523} \\
      \midrule
      $\mathrm{F}^3\text{Mamba}$ {\textbf{(Ours)}} 
        & Face+Finger & \textbf{8.204} &  \textbf{10.115} &  \textbf{12.383} &  \textbf{0.644} &  \underline{9.360} &  \underline{10.938} &  \textbf{15.059} &  \textbf{0.546} \\
      \bottomrule[1.2pt]
  \end{tabular}
  }  \\ \vspace{0.1cm}
  \small
  MAE = Mean Absolute Error in HR estimation (Beats/Min), RMSE = Root Mean Square Error in HR estimation (Beats/Min), MAPE = Mean Percentage Error (\%), $\rho$ = Pearson Correlation in HR estimation.
\end{table*}

To evaluate our the reliability of our model in different clinical environments, we conducted two cross-dataset experiments: (1) training on the Lab dataset and testing on the Clinic dataset (Lab$\to$Clinic), and (2) training on the Clinic dataset and testing on the Lab dataset (Clinic$\to$Lab). These experiments simulate real-world healthcare scenarios where monitoring systems must work across varied patient populations and recording conditions.

As shown in \autoref{tab:cross}, in the Lab$\to$Clinic setting, our model achieves the best performance across all metrics, with an MAE of 8.204 BPM and a $\rho$ of 0.644. For comparison, the best-performing single-view model, the fingertip-based PhysMamba, achieves an MAE of 8.629 BPM and a $\rho$ of 0.599. The relatively small performance decrease from intra-dataset testing (with an MAE of 7.405 BPM) to cross-dataset testing (with an MAE of 8.204 BPM) highlights the adaptability of our $\mathrm{F}^3\text{Mamba}$ model to new clinical environments—a crucial feature for practical cardiac monitoring systems.

In the Clinic$\to$Lab setting, our model performs well for stability and correlation metrics (best RMSE and $\rho$), while slightly trailing PhysMamba in average error metrics. This pattern reflects several clinical monitoring challenges: the Clinic dataset contains less total monitoring time (24 minutes vs. 195 minutes), limiting training data; elderly cardiovascular patients exhibit more hand tremors affecting video quality; and these patients have difficulty maintaining stable camera contact. While our fusion approach mitigates these issues, the dataset differences still impact performance. These results highlight both the challenges and potential of smartphone-based vital sign monitoring across diverse patient populations.

\subsubsection{Ablation Study on F\textsuperscript{3}Mamba Components}
This section evaluates the importance of various components within the $\mathrm{F}^3\text{Mamba}$ model through a series of ablation studies. We constructed distinct model variants (denoted V0-V5) by adjusting or removing key modules, and subsequently performed a detailed comparative analysis. \autoref{tab:ablation module} presents the results of the ablation experiments. All ablation studies were conducted on the Lab dataset using a three-fold cross-subject-validation protocol to obtain the average results.

\begin{table*}[htbp]
  \centering
  \caption{Ablation Study on various components inside $\mathrm{F}^3\text{Mamba}$}
  \label{tab:ablation module}
  \resizebox{\textwidth}{!}{%
  \begin{tabular}{ccccccccc}  
      \toprule[1.2pt]
      {\textbf{Type}} & {\textbf{Face Stream}} & {\textbf{Finger Stream}} & {\textbf{Fusion Stream}} & {\textbf{CSSM}} & \textbf{MAE} $\downarrow$ & \textbf{MAPE} $\downarrow$ & \textbf{RMSE} $\downarrow$ & \textbf{$\rho$ $\uparrow$} \\
      \midrule[0.8pt]
      V0 & $\checkmark$ & $\times$ & $\times$ & $\times$ & 14.510 & 13.810 & 24.078 & 0.353\\
      V1 & $\times$ & $\checkmark$ & $\times$ & $\times$ & 11.132 & 10.764 & 21.304 & 0.538\\
      V2 & $\checkmark$ & $\checkmark$ & $\times$ & $\times$ & 9.256 & 8.994 & 17.614 & 0.568\\
      V3 & $\checkmark$ & $\checkmark$ & $\times$ & $\checkmark$ & 7.184 & 7.229 & 13.951 & 0.578\\
      V4 & $\checkmark$ & $\checkmark$ & $\checkmark$ & $\times$ & 7.537 & 7.556 & 14.338 & 0.541\\
      V5 & $\checkmark$ & $\checkmark$ & $\checkmark$ & $\checkmark$ & \textbf{6.664} & \textbf{6.859} & \textbf{12.796} & \textbf{0.636}\\
      \bottomrule[1.2pt]
  \end{tabular}}
    \\ \vspace{0.1cm}
  \small
  MAE = Mean Absolute Error in HR estimation (Beats/Min), RMSE = Root Mean Square Error in HR estimation (Beats/Min), MAPE = Mean Percentage Error (\%), $\rho$ = Pearson Correlation in HR estimation.
\end{table*}

\textbf{Impact of View Quantity: Dual-view Outperforms Single-view, with Fingertip View Being More Informative:}
Comparisons between single-view (V0 and V1) and dual-view (V2-V5) variants emphasise the importance of the quantity of views. The fingertip view (V1) consistently outperforms the face view (V0) among single-view variants, and dual-view variants outperform any single-view counterpart. The basic dual-view variant (V2) reduces the MAE by 17\% compared to the best single-view variant (V1). This confirms that multi-view inputs enhance the robustness of physiological signals by leveraging complementary spatiotemporal information, thereby improving performance.

\textbf{Effectiveness of Hierarchical Fusion: Progressive Cross-view Interaction Surpasses Single-stage Fusion:} 
Although fusing dual-view features at the final layer (V2) is better than using a single view, it still results in relatively high MAE of 9.26 BPM, indicating that cross-view features are not sufficiently aligned via one-time fusion.
Conducting multi-stage fusion after each TD-Mamba layer reduces the MAE to 7.54 BPM (18.5\% lower than V2), while retaining the complementary advantages of dual views.
This shows that, through 'early interaction + progressive alignment', hierarchical fusion continuously optimises cross-view information at different feature abstraction levels and suppresses view-specific noise more effectively than one-time fusion only at the final layer.

\textbf{Effectiveness of CSSM: Dynamic cross-stream regulation improves fusion:}
Adding CSSM to single-stage fusion variants (V3 vs. V2) reduces the MAE by 22.4\% (from 9.256 to 7.184 BPM). With hierarchical fusion variants (V5 vs. V4), CSSM  lowers MAE by a further 6.2\% (from 7.537 to 6.664 BPM), confirming its key role in robust multi-scale fusion. From a mechanistic perspective, CSSM adaptively updates the state and projection matrices of one stream using cues from the paired stream. This enables reliability-aware gating when a view is corrupted by motion or illumination.

\subsection{Computational Cost}

\begin{table}[htbp]
  \centering
  \caption{COMPLEXITY COMPARISON}
  \label{tab:complexity}
  \resizebox{0.5\columnwidth}{!}{%
  \begin{tabular}{ccccc}  
  \toprule[1.2pt]
  \thead{Methods} & 
  \thead{Param(M) $\downarrow$} & 
  \thead{FLOPs(G) $\downarrow$} &
  \thead{Storage(MB) $\downarrow$} \\
  \midrule[0.8pt]
      PhysNet & 8.85 & 70.32 & 3.38 \\
      PhysFormer & 73.81 & 38.53 & 12.69 \\
      RhythmFormer & 33.26 & 49.53 & 75.8 \\
      PhysMamba & 7.59 & 60.40 & 2.90 \\
      $\mathrm{F}^3\text{Mamba}$ \textbf{(Ours)} & 13.87 & 113.46 & 5.29 \\
  \bottomrule[1.2pt]
  \end{tabular}
  }
\end{table}

In this section, we compare the computational complexity of our model with existing SOTA methods. The results are presented in \autoref{tab:complexity}. All metrics are calculated under a standardized input dimension of $160\times3\times128\times128$. The number of parameters (Param), floating-point operations (FLOPs), and storage requirements are reported for each method.

Our model, $\mathrm{F}^3\text{Mamba}$, has a total of 13.87 million parameters, which is higher than PhysNet and PhysMamba, but significantly lower than PhysFormer and RhythmFormer. The FLOPs of our model are higher than PhysMamba, due to two factors: the introduction of a new view and the addition of cross-view fusion. However, the performance gains justify the increased computational cost. The storage requirement of our model is 5.29 MB, which is also well-suited for deployment on mobile devices, ensuring practicality and efficiency.

\section{Discussion}

\subsection{Dual-View Video-based Physiology Sensing in Mobile Scenarios}
A central design choice in $\mathrm{M}^3\text{PD}$ is to record facial and fingertip videos \emph{at the same time} from a single smartphone in mobile scenarios. This is not only a “more data is better” decision, but a scenario-driven one. In real mobile use, the two views behave very differently: facial rPPG is convenient, contactless, and works well when the user is looking at the screen, but it is sensitive to head pose, distance, and illumination changes, as illustrated by the motion artifacts in \autoref{fig:collection} (c) and the poor performance of single-view facial methods in \autoref{tab:front_view_deeplearning}; fingertip cPPG, in contrast, usually has a stronger pulsatile component because of the flash-assisted, contact illumination, but it is more vulnerable to hand instability and partial detachment, which are common in older or clinical users (\autoref{fig:collection} (d)). By capturing both views in parallel we can (i) study view-specific failure modes on the \emph{same} heartbeat, (ii) train fusion models that fall back to the remaining view when one signal is degraded, and (iii) quantify how much accuracy actually improves when the two optical channels are combined. The superior performance of our dual-view model, shown in \autoref{tab:intra} and \autoref{tab:cross}, validates this approach. This is particularly valuable for continuous smartphone usage (e.g., reading and messaging), where users naturally expose the front camera but can occasionally place their finger on the rear camera, making opportunistic dual-view monitoring feasible without extra hardware.

\subsection{Including Cardiovascular Disease Patients in the Video Physiology Measurements}
Most existing video-based physiological datasets—e.g., PURE or other webcam-based collections—are dominated by young, healthy volunteers under good perfusion and controlled lighting. As our cross-dataset experiments show (\autoref{tab:front_view_deeplearning}, \autoref{tab:back_view_deeplearning}), the models trained solely on such data often fail in real telemedicine scenarios because cardiovascular patients present exactly the opposite characteristics: arrhythmias, reduced stroke volume, peripheral vasoconstriction, as evidenced by the HRV metrics in \autoref{tab:hrv_comparison}. These factors, combined with difficulties in holding the phone steadily (\autoref{fig:collection}), lead to weak or intermittent optical pulses at the face or fingertip and therefore to systematic HR underestimation or beat omission. By explicitly adding 47 cardiovascular patients into $\mathrm{M}^3\text{PD}$ and keeping their recordings \emph{in} the benchmark, we expose rPPG algorithms to the population that needs remote monitoring the most. This also makes the dataset more trustworthy for clinicians and nurses, since they can verify algorithm behavior on elderly and symptomatic subjects rather than extrapolating from student volunteers.

\subsection{Reliable Video-based Pysiological Sensing with $\mathrm{F}^3\text{Mamba}$ Model} 

Our experimental results underscore the limitations of single-view approaches and highlight the significant benefits of dual-view fusion for robust heart rate monitoring in real-world settings. As shown in \autoref{tab:intra} and \autoref{tab:cross}, single-view models, whether using facial or fingertip videos, struggle to achieve consistent accuracy, particularly in the challenging Clinic dataset. Facial rPPG is highly susceptible to motion artifacts and lighting variations common in handheld use, leading to high error rates. While fingertip cPPG generally performs better due to stronger signal quality, it is still vulnerable to signal loss from unstable finger contact, a common issue with elderly patients. Our ablation study (\autoref{tab:ablation module}) confirms that even a simple fusion of both views outperforms the best single-view model. By integrating complementary information, our full $\mathrm{F}^3\text{Mamba}$ model achieves a 21.9-30.2\% reduction in MAE compared to the best single-view baseline, demonstrating that fusion is essential for overcoming the individual weaknesses of each view and achieving clinically reliable performance.

\subsection{Enable Broader Applications with $\mathrm{M}^3\text{PD}$ Datasets}
Although we position $\mathrm{M}^3\text{PD}$ mainly as an rPPG benchmarking resource, its synchronized dual-view (face–fingertip) videos together with physiological references make it useful well beyond single-task HR estimation. Each short capture can be leveraged in daily smartphone interactions to accumulate longitudinal cardiovascular records \emph{without} requiring 24/7 wearables, which is attractive for older or cardiac patients who cannot tolerate continuous devices. The fingertip (rear-camera) channel further enables pulse-presence and signal-quality checks~\cite{wang2025physdrive}, so that a system can fall back to the more reliable view or trigger an alert when the facial signal is corrupted. In addition, the dataset contains respiration, SpO$_2$, and spot BP labels~\cite{tang2025camera}, making it possible to train multi-task models that (i) estimate HR while conditioning on posture/activity, and (ii) learn harder or sparser targets (e.g., BP trend) from the dual-view video streams by exploiting the natural inter-site PPG delay between the facial and fingertip vascular beds~\cite{shoushan2021noncontact}. In other words, $\mathrm{M}^3\text{PD}$ is not limited to “HR from face,” but can serve as a seed dataset for opportunistic, phone-based cardiopulmonary sensing and for building mobile health applications that combine convenience, redundancy, and clinical relevance.

\subsection{Limitations}

Our study has several limitations that suggest directions for future work. First, while the $\mathrm{M}^3\text{PD}$ dataset includes a valuable clinical cohort of 47 cardiovascular patients, its scale and diversity are still limited. It does not fully represent the wide spectrum of cardiac conditions, and transient events like short atrial fibrillation bursts are sparse~\cite{liu2022vidaf}. The current cohort also has a narrow skin-tone range, which may limit the generalization of models trained on this data, as camera-based rPPG performance is known to be affected by skin pigmentation~\cite{tang2023mmpd,tang2024camera}. At the same time, the ground-truth modalities, while clinically appropriate for short sessions, have their own limitations. The reliance on pulse oximetry and spot-check blood pressure, without continuous ECG, restricts beat-level arrhythmia analysis and the development of continuous BP estimation models. Future work should aim to expand the dataset with a more diverse patient population (in terms of both clinical conditions and skin tones) and incorporate ECG for more precise validation of cardiac rhythms~\cite{wang2023ecg}. Extending the recording duration in longitudinal studies would also be crucial for capturing transient cardiovascular events and assessing long-term trends.

\section{Conclusion}
In this paper, we presented $\mathrm{M}^3\text{PD}$, the first smartphone \emph{dual-view} mobile physiological sensing dataset that simultaneously covers a controlled lab study (13 healthy adults) and, 47 cardiovascular patients in clinical cohort. By capturing handheld operation, device heterogeneity, and disease-related physiological variability, $\mathrm{M}^3\text{PD}$ fills a missing benchmark for realistic mobile rPPG evaluation.
Built on this dataset, we further proposed $\mathrm{F}^3\text{Mamba}$, a facial–fingertip fusion framework that uses TD-Mamba branches and an F-Mamba fusion module to dynamically rely on the cleaner view, which reduces heart-rate error by \textbf{21.9–30.2\%} compared to single-view baselines.
We hope that releasing $\mathrm{M}^3\text{PD}$ together with $\mathrm{F}^3\text{Mamba}$ will encourage the community to move toward multi-view, cross-device, and clinically grounded mobile physiological sensing.

\section{LLM Usage Clarification}
We used ChatGPT 5.0 for grammar checking and language polishing to improve the clarity and readability of the manuscript. We used ChatGPT 5.0 for generating illustrative simulated character images in Figure 1 and Figure 5, to visually support the manuscript.

\bibliographystyle{ACM-Reference-Format}
\bibliography{main}

\end{document}